\documentclass[5p]{elsarticle}
\journal{Mechatronics}
\usepackage{graphicx} 
\usepackage{amsmath} 
\usepackage{amssymb}  
\usepackage[dvipsnames]{xcolor}
\usepackage{hyperref, url} 
\usepackage{wrapfig}

\DeclareFontFamily{U}{msb}{}
\DeclareFontShape{U}{msb}{m}{n}{ <5> <6> <7> <8> <9> gen * msbm
<10> <10.95> <12> <14.4> <17.28> <20.74> <24.88> msbm10}{}
\DeclareSymbolFont{AMSb}{U}{msb}{m}{n}
\DeclareMathSymbol{\Reals}{\mathalpha}{AMSb}{'122}
\DeclareMathSymbol{\Naturals}{\mathalpha}{AMSb}{'116}
\DeclareMathSymbol{\Knumbers}{\mathalpha}{AMSb}{'113}
\DeclareMathSymbol{\Rationals}{\mathalpha}{AMSb}{'121}
\DeclareSymbolFont{AMSb}{U}{msb}{m}{n}

\begin{document}
\begin{frontmatter}



\title{Active Learning in Robotics: A Review of Control Principles\tnoteref{label1}}


\author{Annalisa T. Taylor, Thomas A. Berrueta, and Todd D. Murphey\corref{cor1}\fnref{label2}}
\ead{annalisa.taylor@u.northwestern.edu, tberrueta@u.northwestern.edu, t-murphey@northwestern.edu}
\ead[url]{murpheylab.github.io}
\affiliation{organization={Mechanical Engineering, Northwestern University},
            addressline={2145 Sheridan Rd.}, 
            city={Evanston},
            postcode={60208}, 
            state={Illinois},
            country={United States}}

\begin{abstract}
Active learning is a decision-making process. In both abstract and physical settings, active learning demands both analysis and action. This is a review of active learning in robotics, focusing on methods amenable to the demands of embodied learning systems. Robots must be able to learn efficiently and flexibly through continuous online deployment. This poses a distinct set of control-oriented challenges---one must choose suitable measures as objectives, synthesize real-time control, and produce analyses that guarantee performance and safety with limited knowledge of the environment or robot itself. In this work, we survey the fundamental components of robotic active learning systems. We discuss classes of learning tasks that robots typically encounter, measures with which they gauge the information content of observations, and algorithms for generating action plans. Moreover, we provide a variety of examples---from environmental mapping to nonparametric shape estimation---that highlight the qualitative differences between learning tasks, information measures, and control techniques. We conclude with a discussion of control-oriented open challenges, including safety-constrained learning and distributed learning. 
\end{abstract}


\fntext[label2]{This material is based upon work supported by the United States National Science Foundation under Grant CNS 1837515 and by the United States Army Research Office MURI award W911NF-19-1-0233. Any opinions, findings and conclusions or recommendations expressed in this material are those of the authors and do not necessarily reflect the views of the aforementioned institutions.}

\end{frontmatter}

\section{Introduction}
``Perceptual activity is exploratory, probing, searching; percepts do not simply fall onto sensors as rain falls onto ground. We do not just see, we look.'' (R. Bajcsy in her 1988 paper \emph{Active Perception}~\cite{Bajcsy88}). 
The difference between seeing and looking is the presence of action---seeing is passive and looking is active. Unfortunately, we do not use distinct words for passive learning and active learning, often leading to confusing the two and unintentionally treating ``learning'' as passive learning with active learning as an afterthought. Nevertheless, how we acquire data impacts the quality of learning and what is even possible to learn, indicating that control---both analysis and synthesis---in learning will inevitably be important. More than three decades after Bajcsy's comments, the key elements of how control synthesis and analysis should inform learning remain largely unaddressed, and the vast majority of work in learning still focuses on analysis of passively collected data; this body of work makes up a statistical theory of learning. Still absent is an action-oriented theory of learning---a control theory for learning. How should control synthesis affect learning? What sort of feedback interconnections facilitate learning? 

\begin{figure}
    \centering
    \includegraphics[height=3.0in]{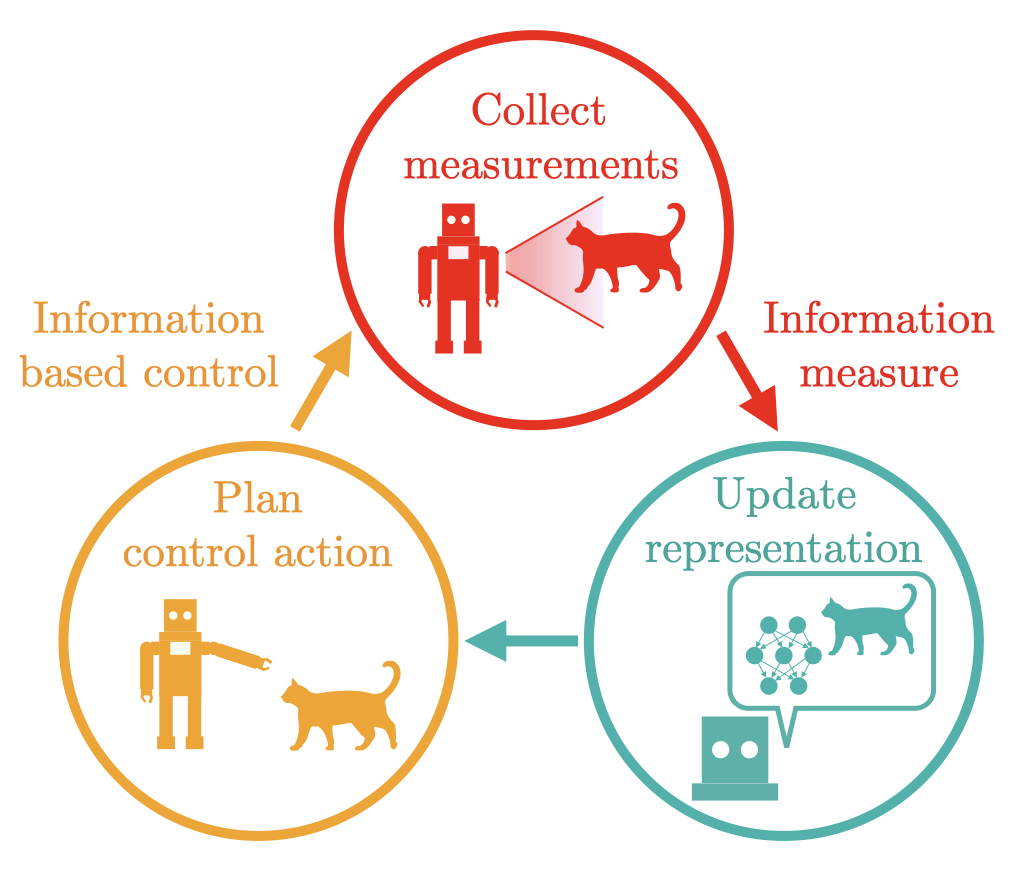}
    \caption{\textbf{The active learning process:} A learner leverages information measures to formulate actions for collecting relevant or descriptive data. Active learning includes the feedback control of a system for which the internal state is both a learning system and history.}
    \label{fig:ALblockdiagram}
\end{figure}

When prior knowledge and existing datasets are widely available, passive learning has proven to be a successful tool for constructing parametric representations of statistical relationships in data. Broadly, passive learning is an optimization process in which the parameters of a model are fit according to data. The last decade has seen major strides in robotics dependent on the advent of modern learning methodologies, particularly variations of deep neural networks~\cite{LeCun2015}. However, in settings where previously existing data sets are unavailable, and where products of human knowledge (\textit{e.g.}, labeled datasets, knowledge graphs) do not exist, a robot will have to engage in unsupervised discovery and acquire the data it needs~\cite{stanton2018situated}. We refer to this process as \textit{active} learning (see Figure~\ref{fig:ALblockdiagram}). In contrast to passive learning, active learning is a decision-making process where agents take actions to gather the data that best realizes a learning objective. 

Animals use their bodies to learn. To paraphrase Bajcsy, we do not just passively learn, we actively learn---the pages of a book do not just turn before our eyes while we absorb information. For agents with physical bodies, such as animals or robots, active learning demands understanding and exploiting the role of embodiment and physical interaction in learning. Insofar as robotics should take inspiration from biology, active learning in robotics will involve the purposeful movement of a robot's body; here, control synthesis tools will connect decision-making to the resulting movement. 

There is a rich literature on how animals use their bodies and movements to improve information acquisition~\cite{martin1965osmotropotaxis,basil2000three,yovel2010optimal,webb2004sensorimotor,khan2012rats,stamper2012active,catania2013stereo,hartmann2001active,nelson2006sensory}. For example, in~\cite{murphey-el2020ChMuMa} we demonstrated that a variety of animals engage in active information acquisition by exploring their environment in proportion to the local amount of perceived uncertainty. In addition to a medium for embodied movement plans, physical bodies are independently capable of implicit computation~\cite{Nakajima2015,Yin2021}, information storage~\cite{Chen2021}, novelty detection~\cite{gold2019selforganized}, and learning~\cite{zhong2020learning}. By harnessing the power of embodiment and morphological computation~\cite{pfeifer2009morphological}, active learning presents a promising way forward for robotics problems where the outcomes of physical interactions may be unknown \textit{a priori}, such as in soft robotics~\cite{rus2015design}.

Not only is embodiment and movement paramount to information acquisition and active learning, but movements themselves can be informative. Recent work analyzing animal and human movement has begun to interpret physical bodies as information channels and motions as information-carrying signals. This has led to the development of methods that help to understand the pathology of conditions such as autism spectrum disorder~\cite{Furutani2021}, schizophrenia~\cite{Osipov2015}, and stroke~\cite{murphey-ral2018BePeFiMu,murphey-sr2018FiAcDeMu} through an information-theoretic analysis of movement. More generally, this suggests that in order to realize learning objectives, active learning requires measures that capture the information content of an agent's movements. 

Counterintuitively, information-rich movement does not always appear productive, orderly, or carefully planned. A well-studied example of this is the optimality of diffusion in animal foraging---here, purely stochastic motion plans have been shown to be highly informative~\cite{Viswanathan1999,Bartumeus2009,Baddeley2019}. Another example of interest to researchers for decades is that of playful behavior in animals~\cite{Bekoff1976,reinhold2019}. One may ask why animals would expend significant energy on movement that is not key to survival; for our purposes, we consider these active behaviors as enhancing learning~\cite{Smith1982}. Hence, to learn through movement, agents must engage in exploratory behaviors that may not always seem useful.

Despite its clear connections to our understanding of learning in animals and humans, the field of active learning finds its origins in theoretical computer science~\cite{Settles2009}. In this setting, agents are represented by disembodied algorithms whose actions are limited to making queries about observed data samples. As a result, many modern frameworks for artificial intelligence have tended to neglect the role of physics and embodiment on the learning process. However, adapting to the constraints of the real world is crucial to learning in the wild. Even the most successful traditional machine learning techniques for robot control, such as reinforcement learning, rely on ``big data'' generated from simulated rollouts. In reality, robot deployment is a time and physically intensive activity, and robots cannot be instantly reset and redeployed at will. To make matters worse, informative data samples are typically sparse. Taken together, these issues highlight the importance of considering sample efficiency and deployment efficiency in robot learning. On the other hand, control theory has a long history of dealing with the constraints imposed by the laws of physics, while simultaneously managing secondary---yet very important---objectives such as safety, robustness, and efficiency.

There are many areas of robotics that will require the type of black-box flexibility of machine learning to make progress. When principled alternatives to modeling the physics of complex interactions between agents and their environments do not exist, machine learning can sometimes be the only way to enable robot control. One area in which flexible learning tools are particularly useful is in high-dimensional nonlinear sensing, where deep convolutional networks excel at integrating potentially hundreds of complex and highly-redundant sensory signals into compressed and informative signals~\cite{Gao2016,Yin2021}. At times, the interactions between a robot and its environment may be infeasible to model either due to properties of the environment (\textit{e.g.}, locomotion in granular media~\cite{Li2013}), or the robot itself (\textit{e.g.}, compliant soft robots~\cite{Laschi2016}). Thus, the field of robotic active learning has the potential to overcome the challenges inherent to robot control and machine learning by inheriting the best qualities of both. In this review, we highlight important progress made towards this goal, and motivate future directions for developing an action-oriented theory of embodied learning.

The organization of this review is as follows. First, we cover the history and basic considerations required for an active learning system---what there is to learn, how to measure information in actions, and how to generate such informative actions. Then, we survey key areas of application for active learning and open challenges in the field. The authors' own work plays a role in creating a narrative, but with consistent reference to the broad literature in robotics on related areas. Section~\ref{sec-history} covers a brief history of the field of active learning and its origins in the broader field of computer science. In Section~\ref{sec-whattolearn} we cover different learning goals, increasing in complexity from learning state parameters to abstract features. Then, Section~\ref{sec-measures} discusses measures of information, focusing on those appropriate to be used as control objectives by synthesis methods such as those in Section~\ref{sec-decisions}. In Section~\ref{sec-applications}, we discuss common applications where facets of active learning naturally arise, whether explicitly or implicitly, in problem formulations. Finally, in Section~\ref{sec-extensions} we discuss extensions and open challenges followed by conclusions in Section~\ref{sec-conclusions}. 

\section{History Of Active Learning} \label{sec-history}
Since its inception, robotics has been interested in making embodied agents learn and adapt to their surroundings like biological organisms~\cite{Merlet2000}. However, due to fundamental limitations on computing hardware, programming machines~\cite{Devol1954} and adaptation to external stimuli~\cite{Walter1951}, robot learning was limited to the most rudimentary demonstrations throughout the mid-20th century.
After establishing his theory of computation~\cite{Turing1937}, Alan Turing shifted his focus to the question of whether machines could think and learn~\cite{Turing1950}. Turing's efforts prompted both the philosophical and formal study of artificial intelligence~\cite{McCarthy1969}.

While hardware posed constraints on applied learning, the second half of the century saw the founding of the field of computational learning theory~\cite{Gold1967,Angluin1980,Valiant1984,Rosenblatt1958}. Analogous to computability theory, computational learning theory focuses on assessing the ``learnability'' of concepts under different models of learning, such as inductive inference~\cite{Angluin1983}, online learning~\cite{Littlestone1988}, statistical query learning~\cite{Kearns1998,Bendavid1995}, among many others. The diversity of models of learning speaks to the difficulty of capturing what we mean when we say that a concept is learnable. To this day, useful models of learning are being introduced to tackle new problems on learnability~\cite{BenDavid2019}. Of the many mathematical frameworks for learning, the most successful and widely used is the Probably Approximately Correct (PAC) learning model~\cite{Valiant1984,Blumer1987,Blumer1989}---a particularly important framework because it was the first to bring insights from the theory of computational complexity to the study of learning. Across its many models of learning, computational learning theory forms the primary means through which we mathematically model and formally understand learning as a computational problem. 

The influence of computability theory~\cite{Cooper2003} is particularly visible in the field's focus on automata theory and linguistics~\cite{Angluin1982}, where problems are often framed as learning languages or equivalent automata specifying the languages. In contrast, much of robotics is grounded in the history of industrial automation, where mechanical interactions are the fundamental object of interest~\cite{Nocks2007}. As a result, robot learning focuses on the role of physics on sensing, actuation, and mechanical interactions with the environment for the purpose of learning.

One of the most important areas within computational learning theory is that of query learning~\cite{Angluin1988,Kearns1998,Cohn1996}. This field is concerned with identifying the classes of functions that a ``learner''  (\textit{e.g.}, an algorithm) can learn by observing samples of data provided by an ``oracle'' (\textit{e.g.}, a teacher or an environment) using a given model of learning. At each stage of the learning process, the learner has a ``learning hypothesis'' about the nature of the function class that it is learning. In the context of query learning, the learner is additionally allowed to ask the oracle for information about the samples it is observing or about its current learning hypothesis~\cite{Settles2009}. The learner then must make decisions about what queries to present to the oracle in order to advance its learning objective~\cite{Feldman2013}. In this way, learning is no longer framed as a passive process. Instead, it is a decision-theoretic process through which the learner takes actions in order to further its objective---or in other words, \textit{active} learning. By leveraging their decision-making, active learners can almost always achieve the same performance as an equivalent passive learner with exponentially fewer data samples~\cite{Balcan2010}. This framing can be restrictive in a robotics context where actions have the potential to elicit information and affect the environment or learning objective. Despite forming a theory grounded in the decision-making of learning agents, computational learning theory has not concerned itself with these types of practical considerations that embodied robot learning demands.

Another theory of learning largely independent from those discussed above is reinforcement learning (RL), which finds its origins in the study of conditioning in psychology~\cite{Watson1913}. As originally envisioned, RL refers to the use of external stimuli and incentive structures to elicit desired behavior out of animals or humans~\cite{Skinner1938}. In this sense, RL was established as a theory of learned behavior rather than learning in-itself. However, its mathematical underpinnings were not established until the second half of the 20th century in the work of Richard Sutton and Andrew Barto among others~\cite{Barto1981,Sutton1981,Barto1983}. By grounding their work in the theory of dynamic programming~\cite{Bellman1966} and optimal control~\cite{Sutton1992}, Sutton and Barto created a rich mathematical theory of learning and control based on the behavioral psychology of reinforcement~\cite{Thorndike1927}. Typically, an RL problem is framed as a Markov Decision Process (MDP) where an agent must take actions in order to explore their environment and learn how to maximize their reward signal~\cite{Sutton2018}. When agents are making decisions and taking actions to actively gather data and learn about their objective, we consider RL to be a type of active learning. In contrast, if exploration is being handled passively through naively randomized simulated experience, we do not.

Despite its early uses for optimal control~\cite{Barto1983}, RL has only recently become a primary technique for robot learning due to the many successes of deep RL in continuous control~\cite{Lillicrap2015,Duan2016,Gu2017}. However, most methods developed for deep RL are ill-suited to robot learning because of their large data requirements, lack of generalizability between tasks, as well as their inability to learn incrementally and guarantee safety~\cite{Kaelbling2020,Levine2021,Sunderhauf2018}. While techniques such as Maximum Entropy RL have taken steps to improve data efficiency and generalizability in robot learning settings~\cite{Haarnoja2017,Haarnoja2018,Eysenbach2021}, deep RL is still far off from seamless deployment in the real world due to its reliance on simulated experience to make progress on learning and control objectives~\cite{Peng2018,Hadsell2017,James2019}. Moreover, easily specifying and incorporating safety~\cite{Garcia2015}, stability~\cite{kolter2019learning}, controllability~\cite{Gehring2013,Tsiamis2021}, or reachability~\cite{Tomlin2014} remains an open challenge. Taken together, these points highlight that---despite being a theory of active learning based on the behavior of embodied agents---RL is underdeveloped for many robotic applications in its present form.

In this section we have briefly outlined the historical development of active learning as a field. Throughout the literature and across its different subfields, we have found that although researchers have had great interest in applying active learning methods to robotics problems, there is still a need for the development of theories of active learning specifically \textit{for} robotics. Such theories of robot learning should center the properties of the agent as an embodied control system with requirements for stability, safety, sample efficiency, and continuous deployment. To this end, much of the work that we present in this review focuses on aspects of embodiment, and suggests the possibility of developing a control-oriented theory of embodied active learning.

\section{What Do Robots Need To Learn?} \label{sec-whattolearn}
\label{sec:representations}
What does a robot need to learn from data? Learning goals can be grouped into problems of increasing sophistication and level of abstraction. Here we will distinguish between learning \textit{parameters}, as a relatively simple starting point, learning \textit{models}, and learning \textit{features}. This division is by no means unique, but provides a useful taxonomy for discussing what learning goals we may have for a robotic system. 

\subsection{Parameters} Learning parameters is relevant in many settings. For instance, one may wish to determine the location of an object, food, or predators. In this case, the parameters of interest are spatial coordinates that localize the object. If the parameters evolve in time (\textit{e.g.}, a mobile object) they may have dynamical properties that can be exploited or learned. If a model is known, parametric filters~\cite{Feder99, Leun06a, Sim05,  VanderHook2012} may be used. When the
posterior probabilities of an inference model are not expected to be approximately Gaussian, nonparametric
filters, such as Bayesian filters~\cite{Marchant2012, Wong05}, histogram
filters~\cite{stachniss2003}, or particle filters~\cite{kreucher2007,Roy2006,lu11} are often used instead. Active learning can be critical to overcoming sensor limitations and identifying a wide variety of parameters. A  salient setting for active learning is near-field sensing. Near-field sensing includes tactile sensing which requires mechanical contact and electrosense, where close proximity is necessary. Hence, when subject to near-field sensing constraints, robots must leverage their agency for successful parameter identification. In far-field sensing, such as cameras and radar at a distance, actions may play a more limited role in parameter identification because the sensor range automatically provides substantial information without the need for movement.

\subsection{Models} Models generalize parameters, and can be models of either the robot itself, such as a model of the dynamics, or the environment, such as a topographical map. The ability of a robot to learn a model of its dynamics is important in rapidly shifting environments where first-principle models struggle to make reliable predictions. The problem of system identification is often parametric, focusing on describing the dynamics using models whose structure and number of parameters are fixed \textit{a priori}, such as in neural networks. However, system identification may be \textit{nonparametric} as well, as in Gaussian process regression and other kernel-based methods. Nonparametric models may be particularly useful when robots operate in unstructured or unknown environments. While parametric models have also been successfully used in this context, it is difficult to know ahead of time that a parametrized model will have the representational capacity to characterize the environment. This has led to the use of models with an increasing number of parameters---sometimes on the order of billions of parameters---to ensure that the network can capture the properties of the environment. 

\subsubsection{Mapping}
Mapping is one form of modeling the environment that emphasizes its geometry. Mapping applications often use occupancy grids~\cite{Bourgault02i,Elfes89}, coverage maps~\cite{stachniss2003}, and Gaussian process regression to represent spatially-varying phenomena or high-dimensional belief spaces~\cite{bender2013,Cao2013,Hoang2014,low2008,Singh2009,souza2014}. 
These techniques presume coverage---that data has been taken over a sufficiently varied area to reconstruct and represent the properties of the environment. The active learning approach instead suggests that an agent reacts to data it collects locally and then adjusts its mapping strategy. While environmental mapping in open air is not an application that necessarily demands the use of active learning methods, other types of environments may not be as straightforward. For example, underwater exploration is difficult because robots are subject to stringent constraints on sensing, actuation, and communication. Here, robots often need to operate in environments where light levels prevent long-range visual monitoring, which demands the use of active learning tools in order to construct motion plans that incrementally adapt to the robot's uncertain measurements~\cite{Picardi2020,Breier2020,Zhang2021}. In~\cite{Fossum2019}, the authors use control and Gaussian process regression to model, map, and actively sample the distribution of phytoplankton in the ocean off the coast of Norway, thereby greatly accelerating environmental monitoring and mapping of oceanic resources.

\begin{figure}[pt!]
    \centering
    \includegraphics[width=3.5in]{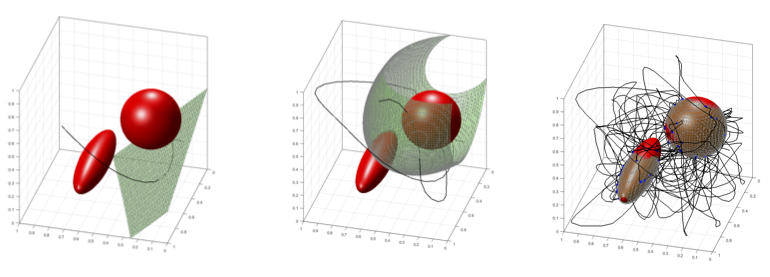}
    \caption{\textbf{Shape reconstruction:} This example  shows the active identification of an unknown geometry in the environment, using binary contact measurements as the measurement modality~\cite{murphey-rss2018AbMaMu}. By developing data-driven models of objects, robots can search for and recognize obstacles or tools without needing analytic or CAD models.}
    \label{fig:shapereconstruction}
\end{figure} 

\subsubsection{Shape}
Similarly to mapping, nonparametric shape estimation is another area of model learning that focuses on the geometric relationship between collected data samples~\cite{Guntuboyina2018}. The shape estimation literature grew from the field of computer vision, and has traditionally focused on static tasks, such as estimating the poses of human bodies~\cite{Hasler2010} or the curvature of roads from image samples~\cite{Taylor2001}. However, as we increasingly deploy autonomy in the real world, determining the shape and material properties of unknown objects may be necessary to interact with them and potentially employ them as tools. To this end, the process of shape estimation may need to be dynamic and probing in nature, requiring that agents leverage their control authority to actively learn the properties of the object. 

As an informative example of this kind of learning problem, we share some results from our own work. In~\cite{murphey-ral2017a}, we considered nonparametric shape estimation using contact-based sensors to actively learn the shapes of obstacles in the robot's environment, which we then extended towards data-driven mapping and localization~\cite{murphey-rss2018AbMaMu}. Figure~\ref{fig:shapereconstruction} shows a three dimensional set of objects whose shapes are being reconstructed from binary contact measurements made by a simulated mobile robot. By actively generating trajectories that make contact with the object surfaces, we maximize the Fisher information of the support vector machine (SVM) object model and successfully identify them. The enabling insight is the use of the Fisher information, which we discuss at length in Section~\ref{sec-measures}, to synthesize object-robot interactions that are optimally informative.

\subsubsection{Dynamics}
One of the most crucial learning tasks is that of identifying the agent's own dynamics. Whether learning the dynamics is necessary due to their intrinsic complexity, or as a result of a sudden malfunction or compliant interaction, there are many scenarios in which it may be impossible or infeasible to have an accurate prior representation of the system's dynamics. Self-identifying dynamics is an active process, where the agent needs to take actions and collect data that explore its different behavioral regimes. In some settings, models that are well-specified in certain behavioral regimes may have to be augmented through data-driven means to work in extraneous conditions. For instance, in aerospace applications data-driven techniques will have excellent data available for nominal conditions but often no data available for specific off-nominal conditions, suggesting the need for active learning outside of the nominal regime~\cite{bayen2007aircraft}. Since the literature on learning dynamics is very diverse, providing a comprehensive survey would require its own review~\cite{Schmidt2009,murphey-tro2019,Berrueta_KoopmanChapter16,Oubbati2010,Nguyen2009}. Instead, here we review a few particular representations of dynamics that are of particular interest to the field of robotics.

Deep neural networks (DNNs) are models comprised of many individual units (\textit{i.e.}, computational synthetic neurons) with limited capabilities that together, through their interconnections, are capable of great representational power~\cite{LeCun2015}. As we have discussed earlier in this review, deep networks are not always suited to the demands of robot learning due to their high data and computational requirements. Nonetheless, certain network architectures have been shown to be well-suited to predicting dynamics, such as recurrent neural networks~\cite{Kim2021}, whose capabilities enable them to predict the global structure of temporal dynamics from local measurements. In settings where learning does not need to occur rapidly or incrementally, carefully chosen deep learning architectures have been successful in learning robot dynamics for control~\cite{Karkus2019,thuruthel2019soft}. While DNNs have been successful in many robotic applications, the online nature of active learning tasks often prevent them from being used in these settings.

A nonparametric alternative to learning dynamics is the use of kernel-based methods~\cite{Hofmann2008}. Kernel regression methods frame learning and estimation problems as one of learning functions embedded in high-dimensional---or even infinite-dimensional---spaces defined over the data domain. The properties of the function space are determined by the choice of kernel, which acts as a generalized inner-product that induces a notion of distance between data samples in the function space. These types of methods have been successfully deployed in robotic systems for both dynamics and inverse dynamics learning~\cite{Schaal2000,Cheng2016,Dalla2019}. However, as typically formulated, kernel methods do not have an easy way to model measurement uncertainty and noise in their function spaces. To this end, Bayesian formulations of kernel methods have been developed~\cite{Smola2003}, the most common of which are Gaussian processes. Gaussian processes (GPs) are one of the primary objects of interest in the study of stochastic processes~\cite{stochasticprocesses}. In GPs, any collection of random variables drawn from the process must be jointly Gaussian. Alternatively, one can insist that functions of the random variables be jointly Gaussian instead, which forms the basis for their application in machine learning~\cite{gaussianprocesses}. In this context, kernels naturally arise in the specification of the mean and covariance statistics of the Gaussian process in function spaces. Using GPs, researchers have been able to parsimoniously incorporate uncertainty and noise into learning robot dynamics~\cite{deisenroth2015}. However, GPs, kernel methods, and nonparametric learning tools at-large typically have difficulty adapting to online learning settings such as robotic active learning. The primary underlying reason is the fact that nonparametric methods tend to grow in complexity as a function of data. Hence, as a robotic agent acquires more data it becomes more computationally expensive to make predictions with the model.

A promising compromise between the representational capacity of neural networks and the simplicity of kernel methods can be found in techniques like the Koopman operator~\cite{Otto2021}. The Koopman operator was first introduced in the study of Hamiltonian dynamics and operator theory~\cite{koopman1931}. Formally, it is an infinite-dimensional, but \textit{linear}, operator that describes the evolution of measure-preserving dynamical systems in a lifted function space. However, to apply Koopman operators numerically they must be approximated in finite dimensions using schemes like Dynamic Mode Decomposition (DMD)~\cite{brunton_dmd,williams_edmd}. Algorithms like DMD use a finite basis for the function space that the Koopman operator acts on to describe the underlying dynamics~\cite{brunton_invariantsubspaces}. Koopman operator theory and its resulting algorithms have been to a large degree developed in the context of dynamics and control, making it an ideal candidate for active learning of dynamics in robotics~\cite{Proctor2018,kaiser2017,murphey-rss2017ToAbMu,bruder2019modeling}. The linearity of the operator lends itself to the use of canonical control techniques such as linear-quadratic regulators, allowing for computationally-efficient nonlinear optimal control~\cite{RSS2019_MamakoukasCastano}. An important feature of this approach is that it does not scale in complexity with data and allows for adaptable incremental learning. The primary caveat with employing these methods is the difficulty of choosing good basis functions with which to describe the dynamics.

\begin{figure}[t!]
    \centering
    \includegraphics[width=0.85\columnwidth]{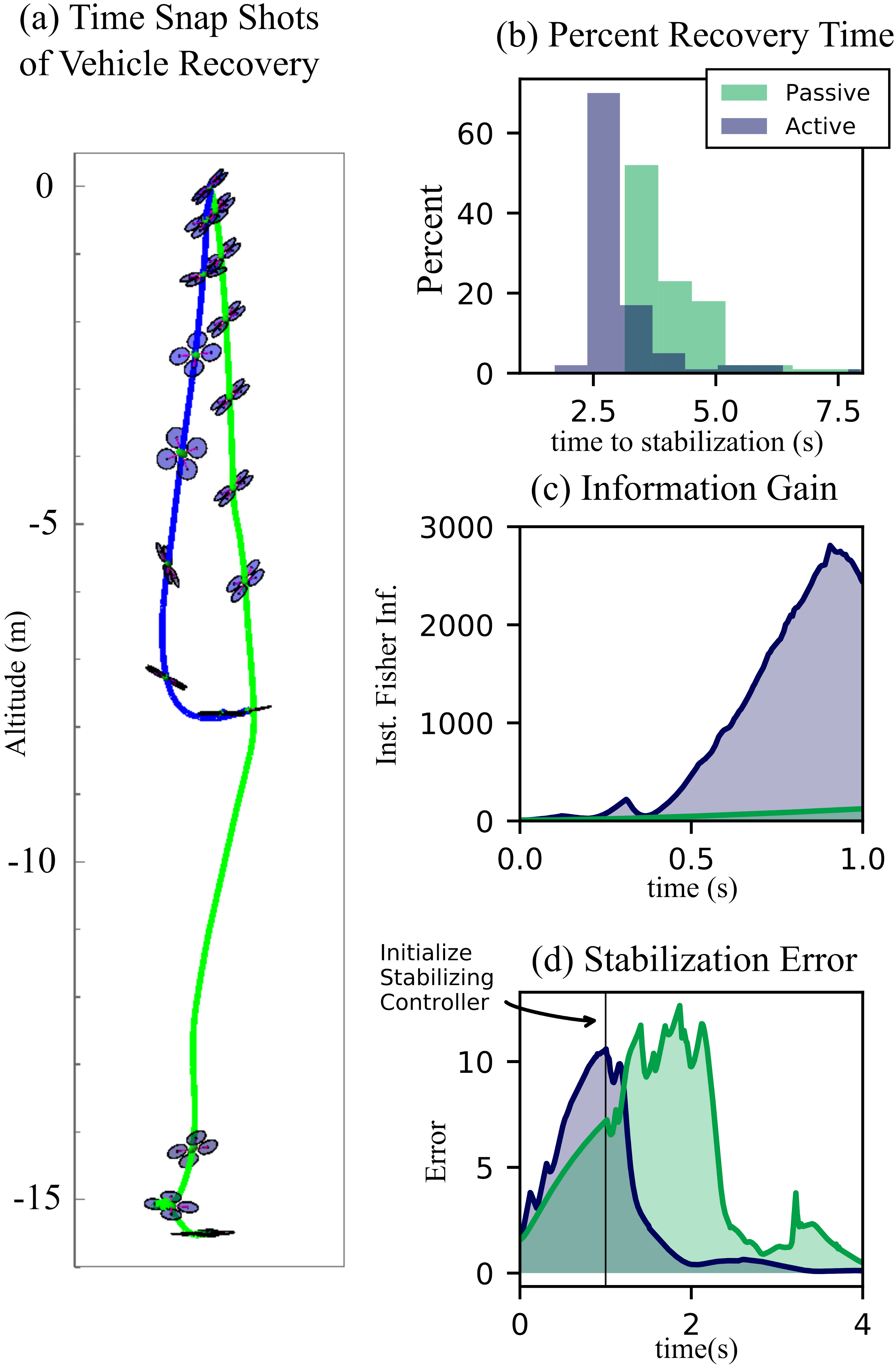}
    \caption{\textbf{Online quadrotor recovery:} Rotor vehicle recovery using active learning in a real-time single-shot learning context~\cite{Berrueta_KoopmanChapter16}. The rotor vehicle with the blue trajectory uses an actively learned Koopman operator representation of its dynamics. The green trajectory is the result of a passively learned Koopman operator representation of the dynamics. The rotor vehicle with the passively learned representation drops further in altitude---potentially crashing---before recovering after one rotor is disabled.}
    \label{fig:dynamicrecovery}
\end{figure}

As an illustrative example of learning dynamics in a context that demands rapid adaptation, we compare passive and active learning in the stabilization of a malfunctioning quadrotor vehicle~\cite{Berrueta_KoopmanChapter16}. In this simulation, we equip two quadrotors with a data-driven model of their nominal dynamics that they can use for model-predictive control. However, at the start of the simulation we disable one of the rotors on each robot causing them to free-fall. To recover, each robot must update their internal dynamics model and stabilize themselves using control. Both agents have a single second during which they can collect data to adapt their dynamics models, after which they switch to a stabilizing controller that tries to regain control of the free-fall. Crucially, one agent learns passively and another actively by optimizing the Fisher information with respect to the unknown Koopman operator, which we discuss in the next section. Figure~\ref{fig:dynamicrecovery}(a) shows snapshots of the different agent trajectories, indicating that the active learning agent is able to stabilize itself much more rapidly than its passive counterpart (see Figure~\ref{fig:dynamicrecovery}(b) as well), potentially avoiding a crash. The active trajectory greatly exceeds the information gain of the passive approach (Figure~\ref{fig:dynamicrecovery}(c)) while also achieving lower stabilization error (Figure~\ref{fig:dynamicrecovery}(d)). Hence, by using control and movement to optimize information measures, robots can learn dynamics faster and more reliably.

\subsection{Features} 
The final and most broad category of learning goals we discuss is that of feature learning. Consider being blindfolded and handed a baseball and a tennis ball in either hand at random. Most people would likely be able to tell them apart with ease. But what is it about either ball that differentiates one from the other? What kinds of properties best represent each ball and its characteristics? Despite having tens of thousands of nerve endings embedded in the palm of our hand, we only need to track a few properties to be able to distinguish between the balls, such as texture and weight. We refer to the general problem of finding informative representations of high-dimensional data that can aid in a task as feature learning.

In the pattern recognition and machine learning literature, features are any measurable characteristics of a phenomenon being observed~\cite{bishop_ml}. Traditional feature learning is the use of machine learning techniques to represent the ``intrinsic'' structure of data from raw and possibly highly-redundant measurements~\cite{Zhong2016}. Recent work in this domain has focused on the use of deep learning towards finding succinct representations of human movement~\cite{Butepage2017} and speech~\cite{Liu2019_speech}. In robotics, tasks are not always well-specified and disentangling the relationship between a robot's internal state and the intended goal may be difficult. This is primarily a challenge in deep reinforcement learning where problems can become intractable when a naive state representation is used. To this end, feature learning can be leveraged towards making deep RL methods computationally tractable, and to develop schemes that better generalize to the variety of sensory inputs to which an RL agent may be exposed~\cite{deBruin2018}. 

A simple example of feature learning can be seen in the Koopman literature. As we previously mentioned, finding the correct choice of basis functions for arbitrary dynamical systems can be very difficult. Nonetheless, recent work has been able to construct basis functions that best describe dynamics---also known as the Koopman operator eigenfunctions, or the intrinsic coordinates of the system---using deep learning~\cite{Lusch2018,kaiser2017,murphey-tro2019}. In general, feature learning of this sort will be particularly important for robots with high-dimensional sensing modalities such as e-skins~\cite{Shih2020}, or computer vision~\cite{Madokoro2014}, and active learning can aid in enhancing rapid identification of intrinsic coordinates.

Our discussion in this review focuses on measures in Section~\ref{sec-measures} and synthesis tools in Section~\ref{sec-decisions} for active learning using location and other low dimensional learning goals as examples.  But the learning goal can be very high dimensional, as in the case learning dynamics of a vehicle, or in the case of learning representations (\textit{e.g.}, machine vision applications). Regardless of whether a learning goal is low dimensional or high dimensional, the robot still has the same control authority to affect learning---it can move its body and take other physical actions to evoke response and facilitate model updates.

\section{Measures for Learning} \label{sec-measures}
Active learning is rooted in the extraction of information from sensors~\cite{Bajcsy88, spletzer03,lu11, dasgupta2006, zhang09, hager91,Benet02,Denzler03}. Accordingly, measures of information should be expected to play a significant role. The aspects of the objective that can be captured by different information measures as well as how this information can be quantified is key in both control analysis and optimal control synthesis. The approach we discuss here follows this perspective, looking for measures appropriate both for information needs and suitable for numerical synthesis. In this section, we cover three important measures relevant to active learning---entropy, Fisher information, and ergodicity.   

\subsection{Entropy}

Entropy-based measures have been employed in a wide range of action sensing results to calculate the expected information gain for each potential action before collecting measurements~\cite{infotaxis,Fox98, Arbel99, vazquez2001, takeuchi1998,Leun06a, kreucher05s,Toh06, Denzler02, zhang09, Tisd09a, Grocholsky06, Singh2009, Roy2006, lu2014,zhang09B, hollinger2013,liao04, emery1998, Ucinski1999, Ucinski2000,Frie04, emery1998, liao04}. This modern concept of entropy was developed by Claude Shannon for use in the communication and transmission of information~\cite{Shannon1948}. Shannon was concerned with the amount of information necessary to reproduce the content of an information source. To this end, entropy is the expected amount of information or “uncertainty” contained in a random variable. In the case of a discrete random variable $X$ where each $x_i$ is a different outcome of the variable, the amount of information content in a particular event is defined by $I(x_i) = -\log p(x_i)$, referred to as bits when in base 2. The entropy of $X$ is the expected value of the information content of each of the possible events.

\begin{equation}\label{entropy}
  H(X)=-\sum_{i}^{}p(x_i) \log p(x_i)
\end{equation}

The information content of a particular event decreases as the probability of that outcome increases, so low probability events provide more information than high probability events. As entropy is the average value of the information content of a random variable, the maximum value of $H(X)$ for $X$, would occur when each outcome of the random variable is equiprobable, \textit{i.e.}, when there is maximum uncertainty about a particular outcome. Thus, any particular outcome for a uniformly distributed random variable does not provide much information. In the context of robotics, this is an explanation for why \emph{rare} or \emph{sparse} events are particularly valuable to a robot's estimation process.

By calculating the Expected Entropy Reduction (EER) of each candidate action, measures of entropy can be readily applied in the context of active sensing. However, exhaustively searching for an optimally informative solution over sensor state space and belief state is a computationally prohibitive process, as it is necessary to calculate an expectation over both the belief and the set of candidate control actions~\cite{Leun06a,Tisd09a,Singh2009,Atanasov2014,kreucher05s, Feder99}. Alternatively, the expected information gain can be locally optimized by selecting a control action based on a local estimate of the expected information~\cite{Grocholsky06, kreucher2007, lu11, Bourgault02i,vazquez2001, Li05, liao04, Wong05, VanderHook2012}. Often times, these methods do not or cannot incorporate general sensor dynamics~\cite{vazquez2001, Li05, liao04, Wong05, VanderHook2012} and even the global strategies are likely to suffer when uncertainty is high and information diffuse~\cite{Rahimi2005, stachniss2003, low2008}.

\subsection{Fisher Information}
\label{sec:VarianceandFI}

Active learning relies on collecting informative sensor measurements to support the learning process. In order to do so, there must be a way to locally measure the information contained in sensor readings. Used commonly in maximum likelihood estimation, Fisher information is a method of quantifying the amount of information that a random variable $X$ contains about an estimate of an unknown parameter, or vector of parameters $\alpha \in \mathbb{R}^M$. Using $p(x|\alpha)$, the density function of $X$ parametrized by the value of the vector $\alpha$, one can determine the likelihood of an observation $x$ given a value of $\alpha$. The Fisher information is an $M \times M$ matrix that captures the local sensitivity between parameters and observations~\cite{cover2012elements,Frie04}:
\begin{equation}
    \mathcal{I}(\alpha)=E_{X}\Big[\Big(\frac{\partial}{\partial \alpha} \log p(x | \alpha)\Big)\Big(\frac{\partial}{\partial \alpha} \log p(x | \alpha)\Big)^\top\Big|\alpha\Big],
\end{equation}
where the expectation is taken over realizations of $X$ at a given value of the parameter vector $\alpha$. If $p(x|\alpha)$ is highly sensitive to changes in $\alpha$---\textit{e.g.}, the distribution of observations exhibits a steep dependence on $\alpha$---then for a given measurement there will be a small range of highly probable values of $\alpha$. If $p(x|\alpha)$ is not sensitive to changes in $\alpha$, then there will be many candidates of comparable likelihood. 

In robotics, Fisher information is well suited  for measurement models that are naturally parametric (\textit{e.g.}, size, weight, location). Measurement models, sometimes called observation models, are predictions of how unknown variables will impact a sensor reading. This sensor reading can be very sophisticated, like a camera being used in a pixels-to-torque application~\cite{wahlstrom2015pixels}, or very simple, such as a one-bit sensor being used for trajectory tracking~\cite{Tovar06,murphey-icra2012c}. The measurement model provides a way of expressing \emph{what} the robot is attempting to learn in terms of its sensing capability and means to adjust its sensors. A commonly used measurement model form is $z=\Upsilon(\alpha,x)+\Delta$, where $z$ is the measurement, $\alpha$ is the parameter being estimated, $x$ is the state of the agent, and $\Delta$ is (possibly multi-dimensional) zero-mean Gaussian noise. This model is in the form of a sum of a deterministic term---typically modeled by first-principle physics---and a noise term which can be rather challenging to justify, since most robotic applications will not have such convenient additive normal distributions. 

\begin{figure}[t!]
    \centering
    \includegraphics[width=3.5in]{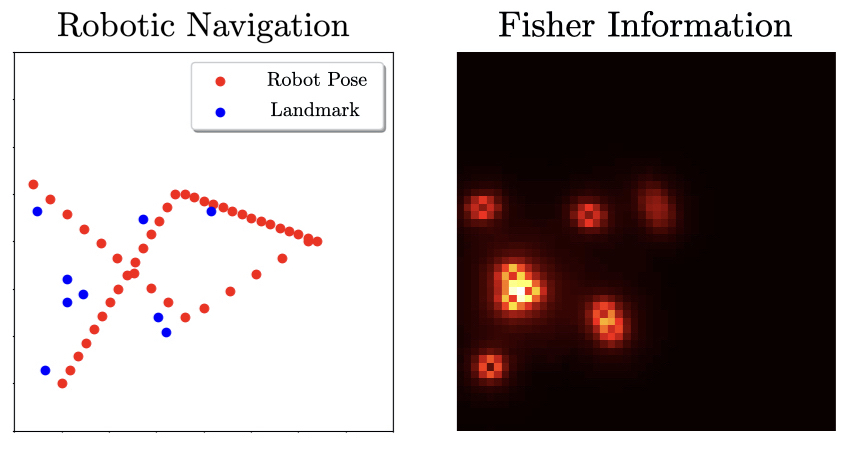}
    \caption{\textbf{Fisher information:} A measurement model indicates how a sensor will respond to the unknown parameter $\alpha$, based on the current state. In SLAM applications, the measurement model might be a model of the detection of landmarks in an environment. The Fisher information distribution over landmarks provides a mechanism for determining what states the dynamical system should achieve to provide maximally informative measurements.}
    \label{fig:MeasModel}
\end{figure}

For active learning applications, measurement models can play an important role in calculating information measures over a space. To estimate a parameter vector $\alpha$, the Fisher information matrix has each element $(i,j)$ given by: 
\begin{equation}
    \mathcal{I}_{i,j}(x,\alpha)=\frac{\partial \Upsilon(\alpha,x)}{\partial \alpha_{i}}^\top\Sigma^{-1} \frac{\partial \Upsilon(\alpha,x)}{\partial \alpha_{j}},
\end{equation}
where the multi-dimensional noise is assumed to be zero-mean Gaussian with covariance $\Sigma$. Intuitively, Fisher information can be expected to be higher where the expected measurement signal is greater than that of the noise. The expected information density $EID(x)$ over a search space can be constructed by computing the \textit{expected} Fisher information with respect to a probability distribution representing an estimate of a parameter $p(\alpha)$. This $EID(x)$ would then form the information landscape against which active learning decisions are made and then executed.

As an example, we consider the use of the Fisher information in Simultaneous Localization and Mapping (SLAM) problems subject to measurement models of the form discussed above. While the SLAM literature in robotics is diverse and well-established, the more recent field of \textit{active} SLAM has seen much growth~\cite{cadena2016past}. Active SLAM makes use of representations of uncertainty and information to generate exploration plans.
In active SLAM, different information measures can capture different features of an environment. In Figure~\ref{fig:MeasModel}, measurement models for landmark detection are used to provide a basis for calculating information measures to inform the agent's exploration plan. In this case, the Fisher information over each landmark attracts the robot to landmarks with lower uncertainty, thereby enabling efficient loop closure. This allows an agent to discern an ensemble of locations that are expected to provide more informative measurements.

\begin{figure}[pt!]
    \centering
    \includegraphics[height=2.5in]{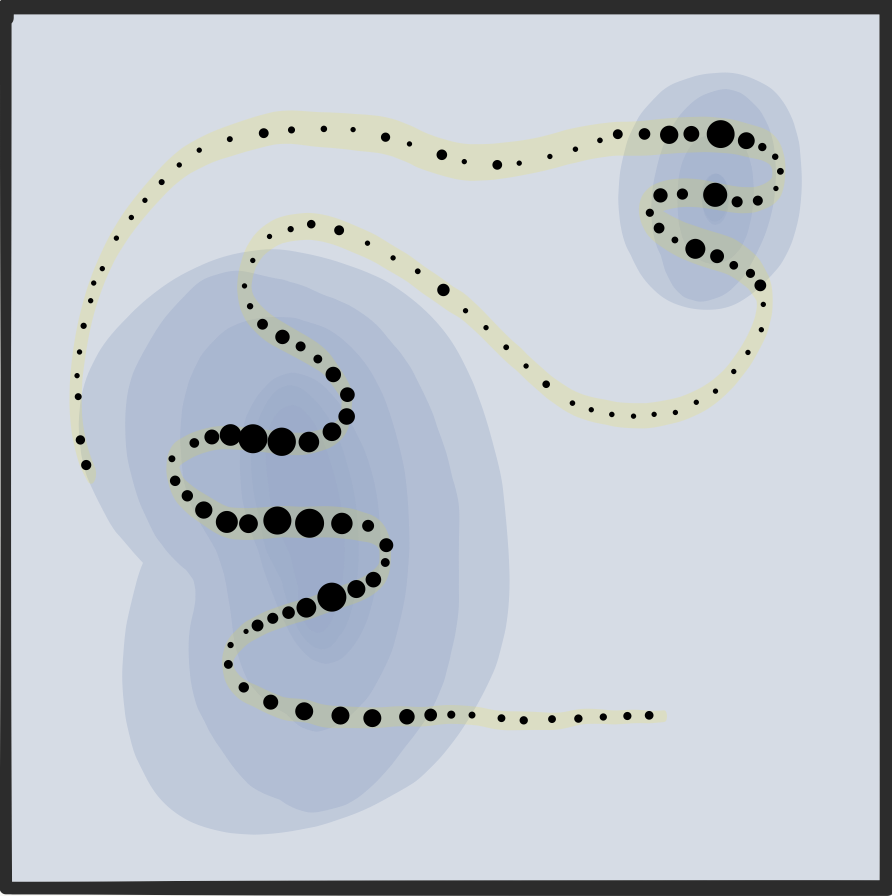}
    \caption{\textbf{Ergodicity:} For an agent to be ergodic with respect to a target distribution, the spatial statistics of the agent's trajectory must match the statistics of the target distribution. This means that the time spent in a particular area is proportional to the density of the target distribution in that area. Here an agent traverses a bimodal distribution. The size of each waypoint is proportional to the time spent at that location.}
    \label{fig:ergodicity}
\end{figure}

\subsection{Ergodicity}
\label{sec:ergodicmeasures}
Ergodicity is a fundamental property of dynamical systems and stochastic processes. Formally, achieving ergodicity implies that the dynamical system uniformly visits all parts of the space in which it exists~\cite{Neumann1932}. However, more often what we mean when we say that a system is ``ergodic'' is whether or not it satisfies the pointwise ergodic theorem~\cite{krengel1985ergodic}. In this sense, being ergodic requires that the system spend time in regions of space in proportion to the measure of said regions. The specific measure used can vary with context, but very often probability measures are used. 

In engineering contexts such as active learning, we are free to choose or construct the spatial measure. Particularly, when a system is ergodic with respect to measures representing an information distribution over the space, ergodicity demands perfect asymptotic sampling of informative states. As a simple example, consider a system trajectory $x(t)\in\mathcal{X}$ and a probability density function (PDF) capturing the expected distribution of information over the space. If the trajectory is ergodic, then the amount of time the agent spends in each neighborhood $\mathcal{N}\subset\mathcal{X}$ is going to be proportional to the amount of information in $\mathcal{N}$ as measured by the PDF (see Figure~\ref{fig:ergodicity}). Hence, designing ergodic dynamics with respect to desired measures is of interest to active learning~\cite{shellMult05ergodic}. However, in order to do so we need a metric that captures how ``ergodic'' our trajectories are. 

Because perfect ergodicity is only possible on infinite time horizons, we require a metric that can be maximized over finite-horizons through decision-making---such a metric was developed in~\cite{Mezic2011}. Metrics on ergodicity provide a principle of motion~\cite{murphey-sr2018FiAcDeMu,murphey-el2020ChMuMa} similar to energy minimization and error minimization, and can be used to synthesize automated exploration for learning, as we will see in Section~\ref{sec-decisions}. The ergodic metric in~\cite{Mezic2011} provides a method for comparing a trajectory $x(t)$---a singleton at any given time $t$---to a distribution $\Phi(x)$ through their \textit{spatial Fourier transforms}. This suggests that one can compare the coefficients $c_k$ of $x(t)$ and $\phi_k$ of $\Phi(x)$ respectively and measure a distance between the two. In general, it is not obvious how one might do this otherwise since information content between dimensionally different objects is typically not well defined. 

\begin{figure}[pt!]
    \centering
    \includegraphics[height=2.25in]{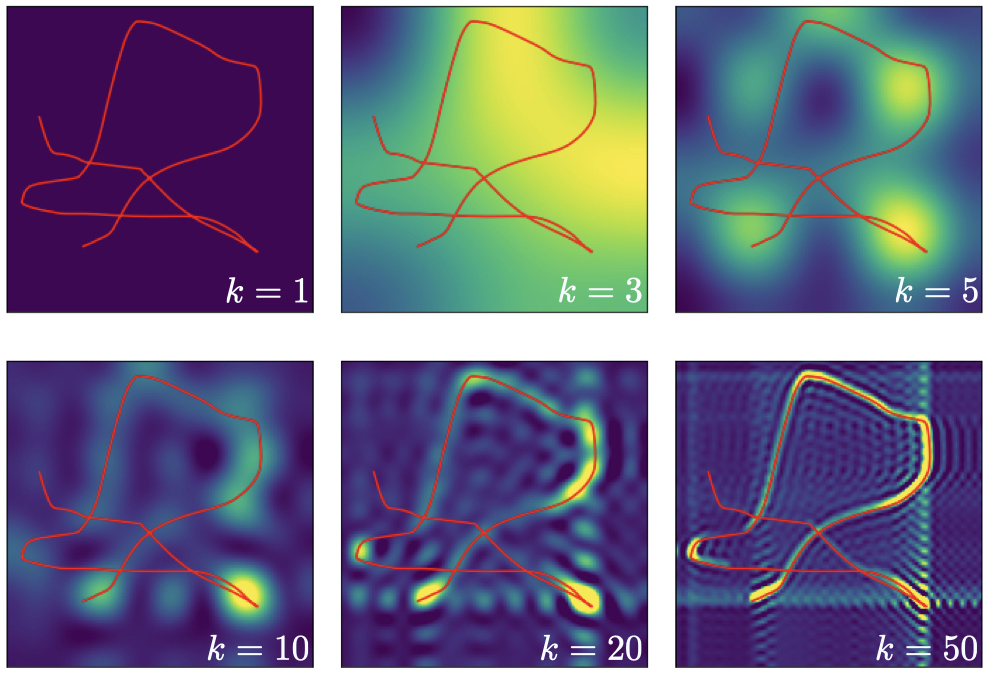}
    \caption{\textbf{Fourier transform of a trajectory:} The Fourier transform of a constant speed trajectory represents the trajectory in the form of a spatial distribution. The representation of the trajectory by its transform changes in granularity for $k=1,3,5,10,20,50$ Fourier coefficients.}
    \label{fig:Fouriercoefficients}
\end{figure}

Comparing how ``close'' two quantities are to each other is imperative for control when using optimization-based methods. To compute the Fourier coefficients $\phi_{ k}$ of a distribution $\Phi( x)$, we use the inner product 
\begin{equation}
    \phi_{k}=\int_{X}\phi(x)F_{k}(x)dx,
\end{equation}
where $F_k$'s represent the choice of Fourier basis functions. For trajectories, we begin by interpreting them as distributions comprised of sequences of impulses:
\begin{equation}\label{eq:timeavedist}
    C(x)=\frac{1}{T}\int_0^T \delta\left[x- x(t))\right]dt,
\end{equation} 
where $\delta$ is the Dirac delta~\cite{Mezic2011}. Then from the properties of the Dirac delta function, we can calculate the Fourier coefficients 
\begin{equation}\label{ck2}
c_{k}=\frac{1}{T}\int_{0}^{T}F_{k}( x(t))dt,
\end{equation}
where the coefficients take on the value of the basis functions averaged over a time window of duration $T$. An example of such a spatial representation is shown in Figure~\ref{fig:Fouriercoefficients}, where a trajectory along with its Fourier decomposition is shown for different numbers of coefficients $c_k$. As the number of coefficients $k$ increases the spatial resolution of the trajectory improves, showing how the statistics of a trajectory may be represented as a spatial distribution. 

With this in mind, the ergodic metric represents a \emph{distance from ergodicity} that is measured from a time-averaged trajectory $x(t)$ with respect to a distribution $\Phi(x)$. This distance is calculated by imposing a norm on the difference between the trajectory's $c_k$ and the distribution's $\phi_k$ coefficients. Particularly, we take the Sobolev norm between the coefficients by using the sum of the weighted squared distance between them:
\begin{equation}\label{ephi}
  \mathcal{E}(x(t))=\sum_{k_1=0}^{K} \cdots \sum_{k_n=0}^{K}\Lambda_k\left|c_{k}-\phi_{k}\right|^2 
\end{equation}
where $K$ is number of Fourier coefficients used for each of the $n$
dimensions, and ${k}$ is a multi-index $(k_1,...,k_n)$. The coefficient  $\Lambda_{k} = (1+||{ k}||^2)^{-s}$ is a weight where $s=\frac{n+1}{2}$, which places larger weight on lower frequency information, ensuring convergence~\cite{Mezic2011}. It is worth noting that spectral methods, and the ability to generate a norm on a trajectory $x(t)$ using them, offer opportunities in measuring entropy as well. The entropy of a distribution could also be measured in the Fourier domain---yielding an objective function that is differentiable and amenable to control synthesis, enabling one to avoid approximating entropy in an optimization with Fisher information (e.g., as in \cite{murphey-tro2014a}).

The measures discussed in this section form the basis for how we measure performance of an active learning system. The next section focuses on synthesizing decisions that optimize, or at least improve, those measures.  

\section{Control Synthesis for Active Learning}
\label{sec-decisions}
Active learning has a wide range of applications in robotics including prioritized decision making~\cite{Cooper08,Toh06}, inspection~\cite{hollinger2013}, mine detection~\cite{Cai2009}, object recognition or classification~\cite{Arbel99, Denzler02, Ye99}, next-best-view problems~\cite{vazquez2001, massios1998, takeuchi1998}, and environmental modeling~\cite{bender2013,Cao2013,marchant2014}. As a result, particular controller architectures may be advantageous for different environments, tasks, and constraints. Here, we survey several model-based optimal control methods that provide distinct advantages for active learning. 

Model-predictive control (MPC) is an optimal control framework that optimizes current actions with respect to an objective while taking into account the future behavior of the system over a finite time horizon. Once the current action is taken, MPCs reoptimize from the new starting point and continually plan actions throughout the receding horizon. MPCs are particularly suited to active learning because receding horizon planning lends itself to continuous incremental learning, while simultaneously enabling assessments of the safety and stability of trajectories. In contrast, other optimal control approaches such as the linear-quadratic regulator (LQR) must solve the entire control problem without replanning.

One of the primary optimal control algorithms is Differential Dynamic Programming (DDP)~\cite{Mayne1966}, which is an extension of the seminal work by Bellman~\cite{Bellman54}. DDP is a model-predictive method requiring second derivatives of the dynamics to realize quadratic convergence to the optimal solution. While DDP has fast convergence guarantees, calculating the Hessian of the dynamics can be computationally intractable. If one is willing to forego the fast convergence rate by disregarding the second order terms of the control solution, DDP becomes equivalent to the first order iterative LQR (iLQR) method. DDP and iLQR have both been shown to be effective in the context of robot control in a variety of applications~\cite{tassa2014control}. For example, in~\cite{kumar2016learning} the authors use local trajectory optimization methods in combination with RL to learn policies for dexterous manipulation with a five-fingered robotic hand. In scenarios where the dynamics are known or easily modelled, and their Jacobians and Hessians are inexpensive to compute, DDP and iLQR may be well-suited to active learning applications.

A method that generalizes MPC to both convex and nonconvex objectives is the sampling-based Model Predictive Path Integral (MPPI) control algorithm~\cite{williamsinformation}. In MPPI, Monte Carlo sampled trajectories are used to approximately extremize a free energy objective~\cite{Theodorou2012}. These types of objectives are designed in analogy to thermodynamic free energy from the statistical mechanics literature and can be used to synthesize control~\cite{Kappen2005}. Moreover, the synthesized control actions are formally equivalent to Bellman optimal control without the need for computing derivatives, and their computation can be easily parallelized~\cite{williams2017model}. As a result, MPPI is particularly well-suited for use in learning problems where the dynamics of the agent are non-differentiable or too complex to differentiate in a computationally-efficient way as with neural network models. For example, in~\cite{williamsinformation} the authors use MPPI to learn a neural network model of the dynamics of an auto-rally autonomous race car. However, depending on the structure of the task, generating enough simulated trajectories to sufficiently sample a learning objective may become prohibitive.

Another model-based control synthesis method is Sequential Action Control (SAC), which is inspired by hybrid systems theory~\cite{SAC}. Unlike other MPC techniques, SAC explicitly tries to expend the least control effort possible in generating actions by taking into account the benefits of taking optimal actions as opposed to alternative policies or doing nothing. SAC simultaneously finds the actions that optimize an objective, the best time to apply said actions, and the application duration. Due to its hybrid specification, SAC can naturally handle non-smooth dynamics, and can also be easily wrapped around other controllers to enable more exotic control architectures~\cite{murphey-wafr2020AbBrPiArMu}. In~\cite{murphey-iros2015a}, SAC was used for active parameter estimation with a robotic system. This work uses SAC to control a robot to determine the length of a pendulum by maximizing the Fisher information with respect to the pendulum parameters. The SAC control actions sequence is piecewise continuous, with generally short application durations for any control. This allows a robot to reactively generate motions towards information dense regions. However, like most MPC techniques, it requires having access to the derivatives of the objective and dynamics, which can constrain its usage in learning scenarios as previously discussed.

An important consideration when choosing a controller for active learning is the global characteristics of the search process. Depending on the structure of the learning task, there may be a single optimum that represents the true parameter value that is being estimated. Other learning tasks require that the agent avoid fixating on a single information source and instead visit many sources. We distinguish between these approaches by referring to them as myopic and non-myopic respectively. Myopic learning uses local algorithms that greedily take actions over short horizons that optimize the immediate learning objective. While these methods are prone to getting trapped in local minima, they have lower computational overhead than non-myopic learning methods. Non-myopic approaches plan control actions over long time horizons so as to produce coverage over distinct information sources. These are often used to avoid local minima associated with fixation~\cite{Singh2009, Cao2013, low2008}, and can take advantage of approximate solutions~\cite{Cao2013, low2008, Hoang2014, stachniss2003,Cai2009, zhang09B, Hollinger2014, Sim05, Leun06a, Ryan2010}.

Choosing a mechanism by which one can avoid myopic learning is critical to operating in environments that have unmodeled effects, such as visual occlusion, where the \emph{expected} most informative state may not provide information. For example, a camera taking a picture of a person behind an a piece of furniture does not benefit from multiple pictures taken from the same state. As a result, dynamic coverage of high information density areas can keep a robot collecting good data while acknowledging unmodeled uncertainty effects through decision-making. Taking these factors into account can be critical to the success of the active learning process. Next, we will examine two approaches to active learning and exploration---infotaxis and ergodic control---that take opposing attitudes towards this question.

\begin{figure}[t!]
    \centering
    \includegraphics[width=3.5
    in]{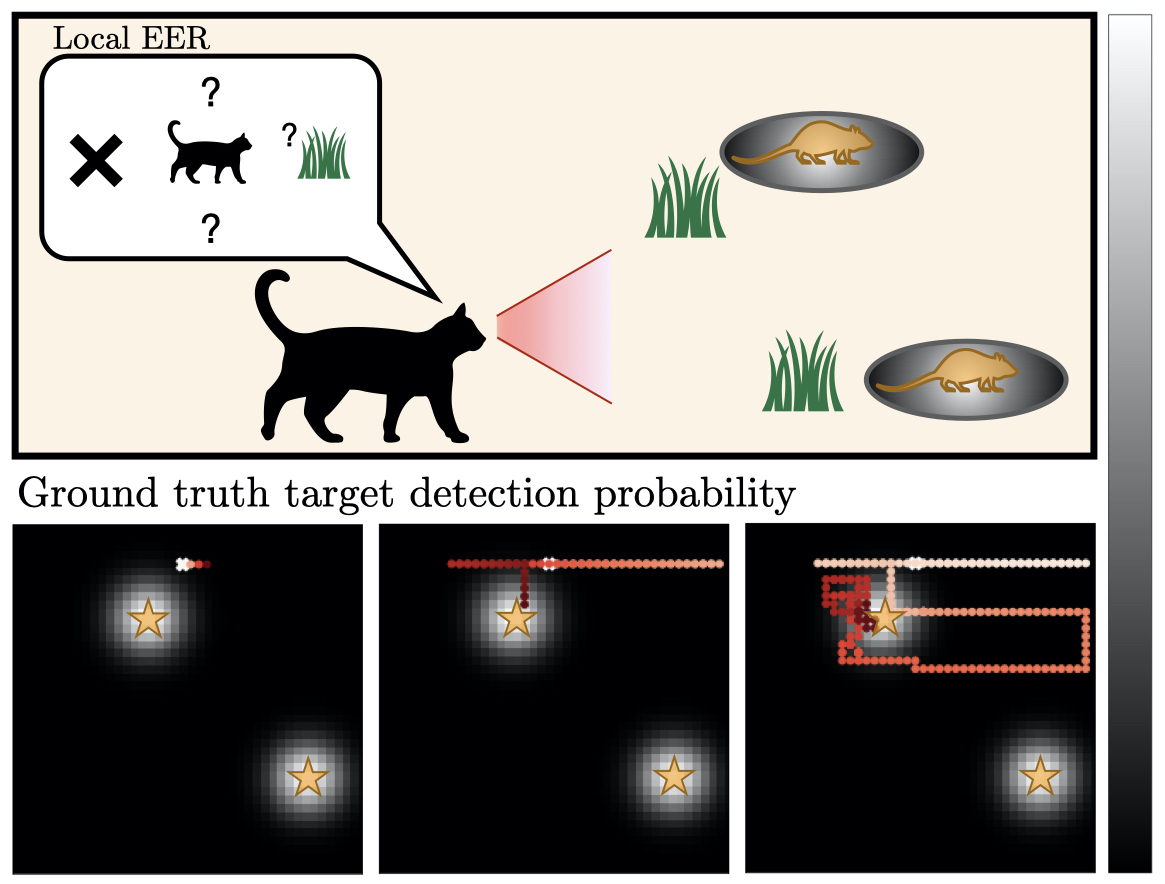}
    \caption{\textbf{Infotaxis:} \textit{Upper panel: } A cat searches for the two yellow mice. The distribution around the mice represents the probability of detecting the mouse at that location where white is high probability. The gradient in the cone around the cat represents the cat's measurement model. The cat has a mental image of the expected local reduction in entropy for moving up, down, left or right. \textit{Lower panel: } This is an example of an infotactic trajectory of an agent searching for target locations, represented by the yellow stars. The likelihood of detecting the target is represented by the distribution around the target locations, becoming more dense closer to the target. The measurement model encodes the ability of a camera to detect an object at a particular range and angle of attack. The search strategy selects the direction of movement that maximizes the expected entropy reduction at each time step. The infotaxis strategy succeeds at finding only one of the two targets in the environment and stops searching.}
    \label{fig:infotaxis}
\end{figure} 

\subsection{Infotaxis}
Inspired by animals' search for chemical sources in a fluid such as air or water, infotaxis is an information-maximizing search strategy using entropy reduction as an information criteria~\cite{infotaxis}. This technique was developed to show that a search plan does not need to depend on environmental gradients, such as the concentration of a scent smoothly increasing in proximity to a flower. Instead, animals may sense traces of a source dispersed by wind or currents and formulate a movement strategy based on infrequent detections.

In this work, an agent attempts to localize a target or source in a 2D environment based on detections of the target. To generate an infotactic trajectory, the searching agent chooses a control action at each time step that locally maximizes the expected reduction in entropy, thereby maximizing expected information gain. Concretely, the agent considers moving to adjacent positions on a lattice, or staying in the same location to take more measurements. 

To determine an action, it is necessary to have a probability distribution $p(r)$ representing the unknown location $r$ of the source. The probability of detecting the source at a given location is dependent on the distance from the source, meaning that the record of detections along the trajectory of the searcher, $x(t)$, carries information about the source location. When a detection event occurs, the times and coordinates are stored in the random variable $\mathcal{T}_t$. From this record of detections, the searcher is able to represent the location of the source as a posterior probability distribution that is updated based on the measurement taken at each  time step.
\begin{equation}
    p_t(r_0) = \frac{\mathcal{L}_{r_0}(\mathcal{T}_t)}{\int \mathcal{L}_{x}(\mathcal{T}_t)dx}
\end{equation}
Here, $\mathcal{L}_{r_0}$ is the likelihood of detections $\mathcal{T}_t$ for a source located at $r_0$. From the posterior distribution one can calculate a control action that minimizes the expected entropy at the next time step by selecting a set of potential actions, computing the EER given the current $p(r)$, and then selecting the action that provides the minimal EER. This strategy can be computationally prohibitive for many systems.

\begin{figure*}[h!]
    \centering
    \includegraphics[width=4.0in]{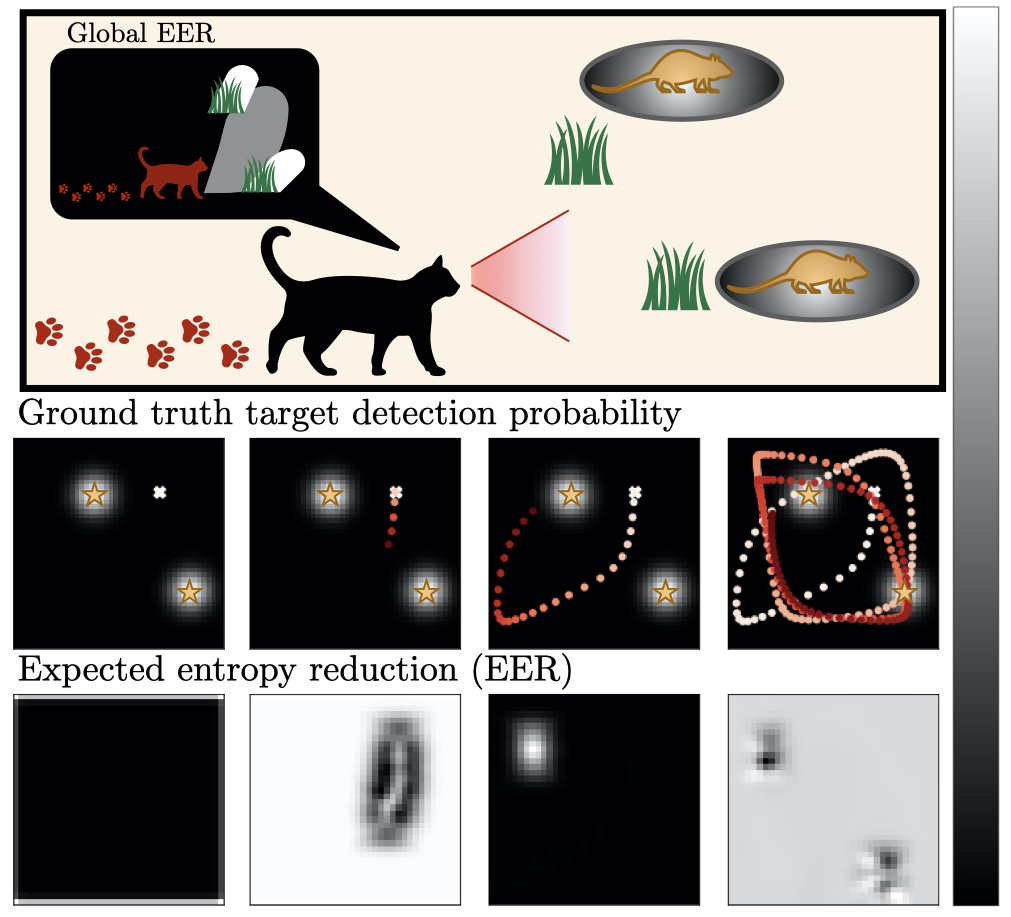}
    \caption{\textbf{Ergodic control with respect to the expected entropy reduction over the search space:} \textit{Upper panel:} A cat searches for two targets represented by the yellow mice. Here, the cat has access to its past trajectory and has a mental image of the expected entropy reduction over the whole search space. \textit{Middle panel:} An agent is searching for two targets located at the yellow stars. The information distribution becomes more dense closer to the targets. Here the agent takes a measurement and updates its belief using the same measurement model and likelihood function as in the infotaxis implementation. \textit{Lower panel:} The agent chooses its next control action based on the global expected entropy reduction. This is determined from its belief of the information content in a particular location.}
    \label{fig:ergodiccontrolexample}
\end{figure*} 
The trajectories produced by infotaxis exhibit similarities to biological organisms such as moths or bacteria that engage in olfactory search~\cite{vickers2000mechanisms}. However, infotaxis-type approaches can fail when there are distractors---states that appear similar to the target but are not the target---in the environment~\cite{murphey-el2020ChMuMa}. The searcher may conflate the actual target with the distractor and then ignore the intended target. Practically, infotaxis can only be implemented using short time horizons as the computational requirements of predicting for longer horizons are significant. For each control action considered, the expected entropy reduction must be calculated, including calculating a posterior for each possible outcome of the measurement random variable. Figure~\ref{fig:infotaxis} provides an example of an infotactic search with two target locations. Here, the agent successfully determines the location of one source and stops searching. This strategy is purposefully ignorant of a signature that may conflict with the perceived location of the target in favor of detecting the same target to increase its certainty. This example illustrates that the infotactic strategy is myopic when confronted with multiple sources or environments with convincing distractors.

While an infotactic search strategy can experience difficulties when there are multiple targets in the environment that require persistent monitoring, it is well suited to react to sporadic cues and requires only local information. Infotaxis represents one of the most straightforward examples of active learning in which an agent acts greedily to maximize expected entropy reduction.

\subsection{Ergodic Control}

Recent work by the authors and colleagues has analyzed biological motion by introducing energy constrained proportional betting \cite{murphey-tro2016,murphey-el2020ChMuMa}, where the energetic cost of movement is balanced against the desire to gain sensory information about a source. This approach uses the ergodic metric, discussed in Section~\ref{sec:ergodicmeasures}, to quantify how well a trajectory covers a distribution of expected entropy reduction. The resulting algorithm produces trajectories that balance informative sampling---collecting many samples in high information areas---with the amount of energy expended from motion. These types of trajectories were observed in the behavior of electric fish, moles, and cockroaches. This suggests that the strategy of energy constrained proportional betting provides a competitive hypothesis for the ways in which living creatures collect information about their surroundings, and may be a robust approach for robotic systems to acquire information. Extensions and variations of this idea now arise in many robotic applications~\cite{nishimura2018sacbp,dressel2019tutorial,dressel2018using,paley2020mobile,chen2020coverage,khodayi2019model,veitch2019ergodic,salman2017multi,ayvali2017ergodic}.

If the goal of infotaxis is to maximize the information content of a series of measurements collected along a trajectory, the goal of ergodic control---first developed in~\cite{Mezic2011}---is to control the spatial statistics of a trajectory $x(t)$ to match those of an expected information density distribution $EID(x)$. This requires the choice of a norm on the difference between the distributions $EID(x)$ and the trajectory $x(t)$ interpreted as a distribution $C(x)$, defined in Equation~\eqref{eq:timeavedist}. To this end, we use the ergodic metric from Section~\ref{sec:ergodicmeasures} as an objective to synthesize maximally ergodic trajectories for general nonlinear systems using tools from model-predictive control~\cite{murphey-tro2016}. However, we note that any trajectory optimization tools or direct optimization tools could be used; we use the results from~\cite{murphey-tro2016} primarily because they are amenable to real-time computation~\cite{murphey-rss2020PrAbScTaPoMu}. 

The first thing to note is that the ergodic metric $\mathcal{E}$ in Equation~\eqref{ephi} is not of the form of a running cost---as a result it is not a Bolza problem (although one can turn it into a Bolza problem by appending the Fourier coefficients to the state vector~\cite{murphey-acc2016b}, creating an infinite dimensional state space). Nevertheless, one can calculate the adjoint variable $\rho$ of the ergodic metric function:
\begin{equation}\label{eq:decentralized_adjoint}
    \dot{\rho} = -\frac{2}{T}\sum_{k } \Lambda_k \left( c_k - \phi_k \right) \frac{\partial F_k}{\partial x} - \frac{\partial f}{\partial x}^\top \rho
\end{equation}
where the dynamics are represented by $\dot{x}=f(x,u)$, and get a descent direction for locally minimizing the ergodic metric~\cite{murphey-tro2018}. Other approaches can be used that lead to slightly different solutions (\textit{e.g.}, the projection-based trajectory optimization method for ergodic control in~\cite{murphey-acc2013c,murphey-case2015,murphey-tro2016}, where higher-order convergence properties come at the expense of high computational cost). A key property of the metric $\mathcal{E}$ is that it is differentiable with respect to $x(t)$, so most optimal control techniques can be easily applied.   

An example of an ergodic trajectory can be seen in Figure~\ref{fig:ergodiccontrolexample}, where the agent is exploring with respect to the expected entropy reduction distribution over the whole environment. The agent is able to successfully locate both target locations in this scenario because the ergodic control strategy is amenable to persistent monitoring of multiple targets. As perfect ergodicity can only be realized as time goes to infinity, the agent will continue to explore the space. Using infotaxis, the agent would conclude its exploration once a target has been detected. Here, we make use of global information to plan control actions over longer time horizons.

With both these local and global information-based synthesis techniques in mind, we next move on to applications in robotics that will depend on active learning strategies. 

\section{Applications in Robotics}
\label{sec-applications}
While the landscape of applications for active learning is almost as broad as that of machine learning itself, here we will focus on settings where datasets are rarely available ahead of time. Active exploration applications such as search and rescue or mapping are particularly relevant in this class of problems, especially when the environments are dynamic and hard to predict. We also discuss applications in which system models are either unknown or difficult to parametrize, as is the case for soft robotics and for many of the areas of application of imitation learning.

\subsection{Soft Robotics}

Soft robots are made from compliant materials, enabling them to be well suited for delicate tasks and environmental adaptation~\cite{agharese2018, galloway2016soft, tolley2014resilient}. Unfortunately, precise modeling and control of soft robots poses challenges because soft materials are continuously deformable and thus nominally have infinite degrees of freedom. There is no clear method of representing the geometry of such a robot without making significant simplifications~\cite{rus2015design}. The most important functional property of a soft robotic system---deformation in response to the environment---makes soft robotic systems practically impossible to meaningfully model for control based on first-principles (\textit{e.g.}, partial differential equations based on elastic body mechanics). Data-driven modeling is a natural alternative when first principle arguments are either not tractable or do not involve the use of a state space.  

Learned representations, such as those constructed by DNNs, have been shown to find input to output mappings that predict the behavior of soft robots~\cite{gillespie2018}. However, these models are difficult to apply using known model-based control techniques. Alternatively, the Koopman operator has also been used for modeling and control of soft robots~\cite{bruder2019modeling}. Described earlier in Section~\ref{sec:representations}, Koopman operators provide a linear representation for nonlinear dynamical systems that is compatible with linear control methods such as LQR synthesis. In practice, a data-driven approximation is adopted. As an example, \cite{bruder2019modeling} develops a model predictive controller with a Koopman operator representation of a soft robotic arm for tracing reference trajectories. The data collection strategy for soft systems plays an important role in determining a model. For instance, though obvious, data collected while an end-effector is out of contact with the environment cannot provide useful modeling data. In prior work we showed that a Koopman operator representation of a robotic system can be actively learned using information-theoretic strategies~\cite{murphey-tro2019}. 

Despite the complexity that soft elastic structures introduce to the analysis of robotic motion, soft robots can beneficially exploit these physical properties. For example, soft structures can be leveraged as a computational resource, sometimes called morphological computation or embodied intelligence~\cite{laschi2014soft}. A soft body that deforms around an object, in principle, will make manipulation easier, and will imply that the amount of explicit computation needed will be lower in exchange for the implicit computation afforded by the soft body. For instance, \cite{picardi2019morphologically} shows that stable hopping behavior of a soft underwater robot can be achieved experimentally by dynamically changing the size of its body. Moreover, with actuator saturation, adapting the morphology of the robot's body was the only route to achieve stable behavior, implying that control over the continuous shape properties of the robot was key to task success.  

In addition to articulation, sensory acquisition via morphological computation is connected to biological systems and present in structures such as the cochlea of the human ear~\cite{mammano1993biophysics} and the bodies of octopuses~\cite{sumbre2005motor}. While data can be passively collected through the physical structure, active sensing is a biologically motivated extension. In~\cite{thuruthel2019soft}, the authors build a perception system to learn the kinematics of a soft actuator and estimate interaction forces  with embedded sensors and recurrent neural networks. In their approach, the authors consider the relationship between action and  perception in the learning process by quantifying sensor information as a result of commanded actuation information. Work in~\cite{sornkarn2016} uses a soft robotic probe to palpate imitation tissue to determine the location of a hard tumor-like nodule. The soft robot was able to adjust its stiffness across iterations of the palpation task based on information metrics calculated from human test subjects. These findings suggest that active haptic perception through physical changes to the probe improves estimation accuracy, motivating active learning techniques that could automate learning for this and other soft systems.

\begin{figure*}[pt!]
    \centering
    \includegraphics[width= 6.0in]{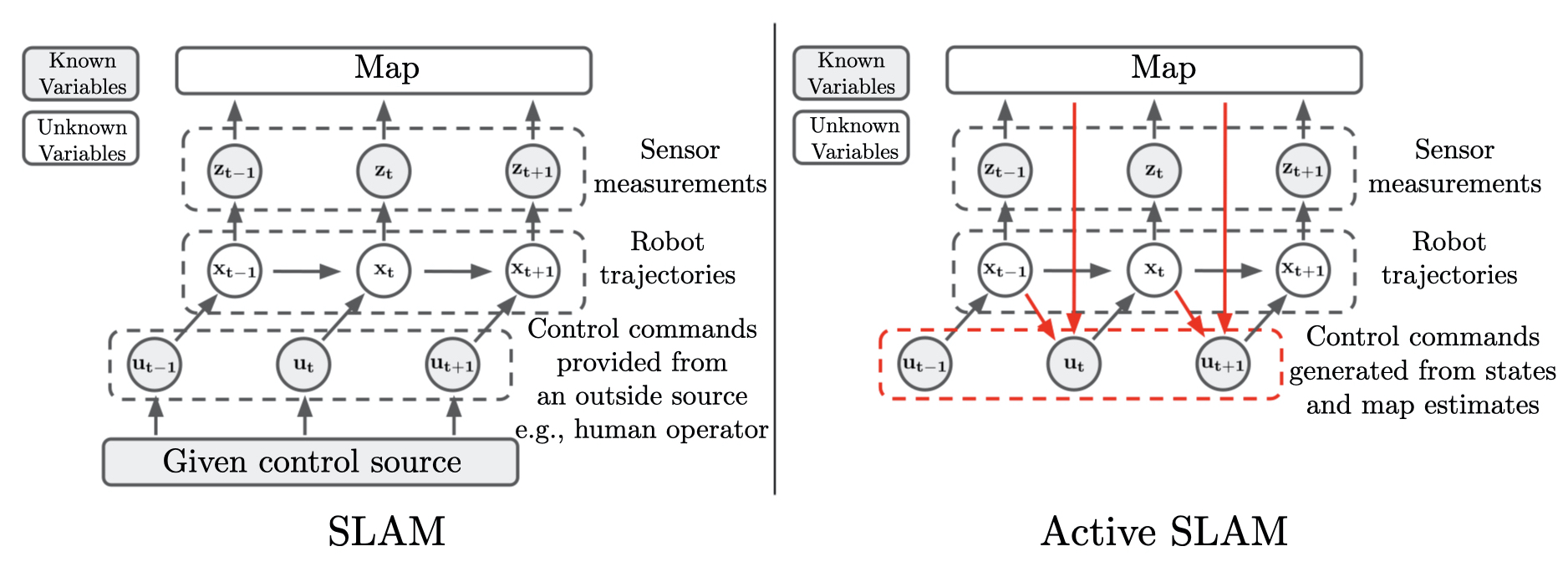}    
    \caption{\textbf{SLAM versus Active SLAM:} Active SLAM uses control commands generated to decrease localization and mapping uncertainty. In traditional SLAM, the control signal is given in the problem statement.}
    \label{fig:SLAM}
\end{figure*}

\subsection{Search and Rescue} 

Prevention, response, and recovery from disasters can be dangerous for emergency professionals who may need to interact with areas affected by events such as hurricanes, oil spills, and earthquakes. Disaster robotics is an area that works to augment the capabilities of workers by delivering real-time data to experts and intervening in the environment~\cite{murphy2016disaster}. The need to efficiently search an environment is an issue at the core of disaster robotics. One of the most visible examples of the need to search an extremely large, dynamic environment in recent years is the investigation of the crash site of Malaysia Airlines Flight 370 (MH370) in March of 2014. In the first 52 days after the crash, the Australian government reported that air crafts and surface vessels covered an area of over 1.6 million square miles. By June of 2018, the final search effort was suspended without success. Although there may be many points of failure in this search effort, one dimension involved robotic technologies that scanned the bottom of the ocean that were incapable of reasoning about potential debris signatures, the dynamic environment, and their own capabilities.

When searching large areas where information is sparse, active coverage algorithms are important in determining important areas of a search region and the schedule to visit these regions. Coverage algorithms are used in many robotic applications such as underwater exploration~\cite{paull2013}, agriculture~\cite{oksanen2009coverage}, and inspection~\cite{englot2012sampling}. The goal of coverage algorithms is to visit all points in an area or volume while avoiding obstacles~\cite{galceran2013survey}. Commonly used approaches for coverage, a taxonomy of which is included in~\cite{choset2001coverage}, include cellular decomposition or grid-based methods to divide the area into manageable sections~\cite{karapetyan2017, jan2014, schwager2009decentralized, stergiopoulos2013spatially}. However, as the complexity of the environment increases, the number of cells necessary to represent the environment increases. These methods typically do not take into account the physical properties of sensing capabilities of the robots or the dynamics of the environment. As a result, coverage is treated as both necessary and sufficient for capturing needed data. This attitude about coverage can be seen in the search strategy of the MH370 investigation which focused on area coverage, neglecting factors such as how the ocean currents might pull debris away from the site~\cite{2015MH370}.

\subsection{Localization and Mapping}

SLAM algorithms create a map of an unknown environment while simultaneously estimating the state of the robot within that environment. This is a major success story in robotics, with the current flood of driverless car technologies all dependent upon SLAM algorithms. When navigating an unknown environment, a robot may lose its ability to localize itself due to accumulated small errors in sensors and actuators, known as representation drift. To correct for this drift, SLAM algorithms use loop closure---the task of identifying whether an agent has returned to a previously visited location---to maintain an accurate representation of the location of the robot relative to environmental features. To maintain loop closure, the robot revisits regions with low estimation uncertainty or informative features to combat representation drift. Beyond localization, loop closure allows the robot to represent the topology of the environment, instead of simply a record of where it has been. 

In passive approaches, a robot performs SLAM with sensor information provided to it. For instance a lidar sensor collects data while driving down a road. In contrast, active SLAM leverages the actions of the robot to seek out informative measurements that efficiently decrease localization and mapping uncertainty. Figure~\ref{fig:SLAM} illustrates the flow of information in passive versus active SLAM. Active SLAM generates controls based on the current state of both the map estimate and robot states. The review paper~\cite{cadena2016past} summarizes methods that have been employed in the development of active SLAM including the theory of optimal experimental design~\cite{rodriguezarevalo2018}, information theoretic approaches~\cite{stachniss2005information, carrillo2015autonomous, carlone2014active} and control theoretic approaches~\cite{leung2006active, atanasov2015decentralized}. Active SLAM can also be formulated as a Partially Observable MDP (POMDP) and approximated using Bayesian optimization or Gaussian belief propagation to attain computational tractability. Belief space planning entails planning in the space of probabilistic estimates of a robot's state and additional variables of interest~\cite{bonet2000, platt2010belief}. This method has also been used in combination with navigation error~\cite{prentice2009belief, valencia2013, patil2015scaling}. 

Using planning algorithms in SLAM is challenging because SLAM is generally executed on a pre-planned trajectory. This trajectory can greatly affect the quality of performance. Conversely, path planning algorithms typically assume a given map. Hence, planning and SLAM are nontrivially interdependent. Work in~\cite{kim2015active} attempts to integrate SLAM with a coverage path planning problem by developing a movement strategy they call perception-driven navigation. The authors use a cost function that weights navigation uncertainty, evaluated using the Fisher information matrix described in Section~\ref{sec-measures}, with the ratio of unexplored to total coverage area. This method plans paths between waypoints that are selected based on a measure of visual saliency, prioritizing areas in which notable environmental features have been detected. The integration of perception based navigation in the SLAM framework is key to balancing effective mapping alongside exploration as the distribution of features in an environment is often highly uneven. It also allows for operating in limited field of view environments, such as underwater inspection tasks.

Developing a method to determine informative features from images is an important aspect of visual SLAM, in which SLAM is performed using only camera inputs~\cite{taketomi2017visual}. Image pre-processing with feature selection reduces the computational burden of scanning  all the pixels in images, leading to many active feature selection algorithms~\cite{chen2011active}. One can also selectively process informative regions of images or videos using a recurrent neural network~\cite{mnih2014recurrent}. Lastly, visual-inertial navigation---where a robot must estimate its state using only a camera and inertial sensors---can supplement the visual SLAM process. In~\cite{carlone2018attention} visual-inertial navigation selects features based on the state of the observer and the context of the scene, using information theoretic constructions as a basis for prioritizing features to be used in state estimation.

\subsection{Imitation Learning}
\label{sec:imitationlearning}

Imitation learning is a widely used and effective method of imparting human skills to robots by learning desired behaviors from demonstrations. To transfer knowledge about a task through imitation, it is important to capture salient features of a demonstration in efficient and generalizable representations of a skill. Here, active learning can play an important role in capturing knowledge from a demonstration.

The field of imitation learning is expansive~\cite{Hussein2017, argall2009survey, osa2018algorithmic} and has been used in numerous settings including autonomous driving~\cite{codevilla2018end}, virtual games~\cite{silver2016mastering}, and replicating human motion in robots~\cite{ijspeert2002}. Capturing knowledge about a task from human experts is especially applicable to robotics, where autonomous systems are charged with operating in complex and unstructured environments. In these situations it can be difficult to manually program specific behaviors and engineer reward functions to suit a task. Imitation learning is commonly tied to deep neural networks to take state/action pairs from demonstrations and learn a policy for a skill. This can often require large amounts of data, leading to questions about what aspects of demonstrations are particularly useful to impart a skill to an autonomous system.

When transferring skills from a human operator to a robot, active learning occurs when a human operator is queried for information. For instance, work in~\cite{silver2012active} considers two approaches to active learning from demonstration in the context of autonomous navigation. A learner, such as a robot, selects expert demonstrations that they believe to be informative based on either novelty or uncertainty reduction criteria. In novelty management, demonstrations are selected based on a density model from which a test feature vector can be compared to demonstrations previously seen in training to provide exposure to unobserved or anomalous data. For uncertainty reduction based active learning, the authors used the Query Bagging Method~\cite{dima2005active}, in which training data is partitioned into multiple subsets. A demonstration would be deemed to have high uncertainty if the variance over these subsets for the demonstration was high.

Inverse reinforcement learning (IRL), also called inverse optimal control, is a method of determining the goals of desired behavior from trajectories executing a policy~\cite{ab2020inverse}. The aim of IRL is to find a reward function that describes the desired task from expert demonstrations. When a task is well suited to be described by a single reward function, IRL is most applicable. However, a policy may be optimal for multiple reward functions, making it difficult to discern intent. In response, it may be necessary to include other objectives. Work in~\cite{daniel2015active, judah2012active} focuses on active learning in the context of IRL, which seeks to reduce the demonstrations from full trajectories to particularly useful states. In this case, active learning means selecting particularly informative samples to be labeled by an oracle. In~\cite{daniel2015active}, a robot learns a reward function and movement policy for a grasping task. The reward function is in the form of a Gaussian process model and is based on human evaluations of the quality of the grasp. In this method, the learning agent is able to impact the demonstrations it sees by choosing to query human expert ratings based on acquisition functions from the Bayesian optimization literature.

Generative adversarial imitation learning (GAIL) is a model-free imitation learning approach that scales well to high dimensional environments~\cite{ho2016generative}. Inspired by generative adversarial networks, GAIL produces behaviors similar to demonstrated behaviors while training a discriminator to differentiate expert attempts with generated attempts. An extension of GAIL, called InfoGAIL, attempts to find latent structure across human demonstrations---that can be highly variable---to describe interpretable concepts~\cite{li2017infogail}. Related to techniques that train a discriminator to differentiate between expert and learned policies(such as InfoGAN~\cite{chen2016infogan}), InfoGAIL approximately maximizes mutual information between latent space and trajectories to deduce meaningful latent variables. In this way, it is possible to produce semantically meaningful or informative data that pertains to a particular task. 

Imitation learning, and the other applications mentioned above, stand to benefit from robots that physically manipulate when and how they learn, rather than relying on visual and aural requests for more or better data, which is one of the principal goals of active learning in robotics. 

\section{Open Challenges} 
\label{sec-extensions}

Closed-loop active learning presents a key opportunity for improving the quality and rate of learning. In this section, we focus on specific challenges in both the near and far term, such as safety and distributability. These challenges are specific to the expertise of the controls community---\textit{e.g.}, analyzing properties like complexity, convergence, and motion feasibility. We end with a broader discussion of questions such as how can one assess the sufficiency of a learning model for a given task? These challenges, among others that we may not yet understand, are at the core of what it means to construct a robotic theory of active learning.

\subsection{Distributability}

Distributability has become a widely studied and often implemented goal for control systems, enabling a swarm of robots to accomplish what an individual robot cannot. In the context of control-driven tasks such as exploration or search, the benefits of distributability are immediately apparent---multiple robots will be able to cover an area more efficiently than a single robot could. Distributed data collection of this kind has been widely and successfully applied in a variety of contexts, such as environmental monitoring~\cite{schwager2009decentralized,low2012decentralized}. The key feature underlying the success of these distributed control applications is that the dynamics of the robot collective are factorable into a block-diagonal representation---the dynamics of each robot agent are independent from one another~\cite{murphey-ral2018AbMu,murphey-rss2020PrAbScTaPoMu}. However, can we expect this to be the case across active learning applications?

While independent robots can easily coordinate to collect measurements and effectively augment their perception~\cite{best2019dec}, learning collectively may prove to be much more challenging for a variety of reasons. For one, when robots are not just collecting data but also using it to learn as a group, they must be in constant communication and sharing data samples with one another. Another important challenge is that the data samples that each agent is locally exposed to may be statistically distinct. Moreover, the noise and disturbances that robots are exposed to may be heterogenous across agents as well. Taken together, these observations suggest that during distributed learning the samples that a swarm collects may not be independent and identically distributed, which is a key assumption underlying most learning methods and can create issues with fundamental properties of the learning process (\textit{e.g.}, convergence). Most of the difficulties outlined so far have been described by the fields of distributed~\cite{Verbraeken2020} and federated~\cite{li2020federated} machine learning. Hence, the success of distributed active learning is in part tied to the challenges of distributed learning generally. 

Nonetheless, some challenges in distributability will be unique to active learning. As we have discussed, when the dynamics of robotic agents are left uncoupled making control decisions may be simple. However, active learning in robotics precisely requires a coupling between learning and taking actions. Then, when agents share a common distributed learning objective, their dynamics may become effectively coupled through the contingent relationship between acting and learning. As a best-case scenario, this can lead to redundant data collection and learning, but in the worst-case this can create stability issues in the learning process. Highly-coupled dynamics, along with extended network dropouts, will generate high degrees of disagreement between agents, making both analysis and prediction more difficult. Thus, eliciting useful collective behavior from decentralized systems based on local decisions is still an open challenge.

\begin{figure}[pt!]
    \centering
    \includegraphics[width=3.5in]{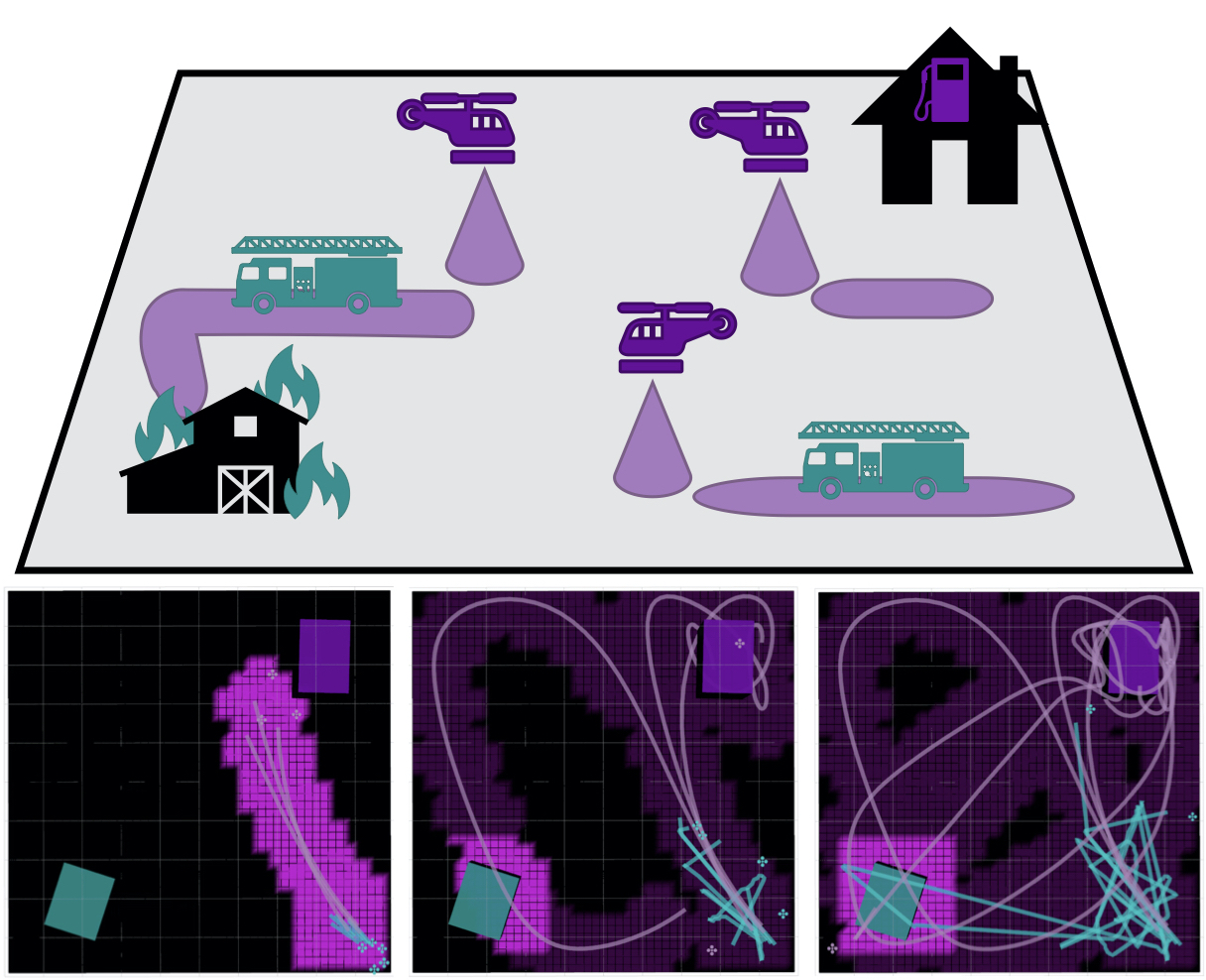}
    \caption{\textbf{Safe decentralized ergodic control:} \textit{Upper panel:} Fire trucks attempt to reach the site of a fire guided by helicopters above. The firetrucks are able to explore in the areas that have already been explore by the helicopters. At the same time, the helicopters must maintain the ability to return to a refueling station. \textit{Lower panel:} Here are three time snapshots of an ensemble of six robots---three purple and three blue---explore an environment subject to the condition that blue robots can only go some place purple robots have already visited. The purple robots are tasked with exploring the purple building while the blue robots are tasked with exploring the blue building.}
    \label{fig:safedecentralized}
\end{figure} 

\subsection{Safe Active Learning}
Safety is a problem of both specification and prediction---one needs to specify what is meant by safety and be able to predict that the specification will be satisfied. Imposing safety enables learning in high-consequence environments with continuous deployment, making reliance on models and prior experience less risky.

Common tools available for imposing safety constraints often depend on Control Lyapunov Functions (CLFs)~\cite{ames2014rapidly,ames2013towards}. These control approaches enforce stability properties of a robotic system through a feedback stabilizing control law that drives a positive-definite differentiable function to zero over time. In the context of active learning, one may desire to have a CLF for ergodic control, using the ergodic metric as the candidate Lyapunov function \cite{murphey-tro2018}. One can use Control Barrier Functions (CBFs)~\cite{ames2019control, ames2016control, wang2017safety} that encode safety constraints, such as in Figure~\ref{fig:safedecentralized} where we impose the constraint that one set of vehicles can only enter a region after another set of vehicles has explored it. Both CLFs and CBFs can be combined with other objective functions that are task-oriented rather than safety oriented; these often then involve solving quadratic programs to satisfy safety constraints~\cite{ames2013towards, ames2016control, berkenkamp2017safe, choi2020reinforcement}. The CLF/CBF approach is the most amenable to computation in high dimensional spaces, but in lower dimensional spaces one can directly solve for safety sets using reachability analysis, which depends on solving a Hamilton-Jacobi-Isaacs partial differential equation~\cite{akametalu2014reachability,bansal2017hamilton}. Though not necessarily practical for high dimensional systems, this guarantees an optimal trade-off between safety and performance.  

An important challenge in these safe learning techniques is that they are model-based. They require a model to evaluate the monotonic decrease of the CLFs/CBFs or to evaluate reachability conditions. Since a robot will typically be learning something about the environment relevant to its evolution, its own dynamics, or its interactions with the environment, all these techniques will rely on model updates of some form along with real-time updates to statistical analysis. A key question is how should a robot stay safe during this process, and what should safety mean when representations critical to safety are not known?

In recent work---following the CLF/CBF viewpoint of safety---we showed that one can use hybrid control methods to schedule switching between a safe controller and a learning controller, while maintaining the asymptotic properties of the safe controller~\cite{murphey-tase2019AbPrMu}. The critical assumption in that work is that there is an operating point where stability of the robot-environment combination is already established and using the safety of that state as a starting point for safe learning. This is often a reasonable assumption; for instance, one might have an empirically safe PID controller for a humanoid robot near upright posture without having model-based safety analysis. Additionally, CBFs have been used to guide the learning process in reinforcement learning~\cite{cheng2019end}. In this work, the CBFs restrict exploration to safe policies and become less conservative as an online learning process learns a model of the dynamical system. This makes the learning process more efficient while guaranteeing safety. This method incorporates online measurements to improve the CBF-RL controller, providing an opportunity for  active learning approaches such as those discussed here to facilitate information gathering. Other approaches to simultaneously satisfying safety guarantees with \emph{a priori} unknown dynamics and/or unknown environments need control formalisms that enforce safety criteria in the absence of any certainty.

\subsection{Stability, Invariance, and Specification}

Another concern critical to learning is how to impose prior knowledge on learned models. Particularly in the context of physical learning, where a model does not need to be an ordinary differential equation or a statistical pattern, but can instead be a principle (such as a motion symmetry~\cite{marsden2013introduction} or energetic dissipation). Among these principled statements of modeling assumptions, stability, the property that the unforced system asymptotically converges to an equilibrium, may be the most common property in a physical system that we may wish to insist upon~\cite{stableModels_Nelson}. In~\cite{murphey-neurips2020}---following~\cite{constraintGeneration_Boots,WLS_stableLDS,kolter2019learning,LyapunovStableFlowPrediction_Erichson,stabilityCertificates_Boffi,richards18a}---we used recent results in  linear algebra to project linear operators (such as  the Koopman representations discussed earlier) onto the closest stable linear operators. Moreover, in~\cite{murphey-tro2020MaAbMu} we applied these techniques to robotic manipulation examples, where notably the experiments were implausible without constraining the learning to stable models. 

There is a wide range of potential specifications one may wish to impose on a learning system. How would one specify that a learned model must satisfy a linear temporal logic (LTL) constraint such as those described in~\cite{mehta2018robot}? What about symmetries in time and space, implying conservation of energy and momentum? Developing formally correct methods for combining learning tools with these specifications is a key step forward towards robot learning under user-generated constraints on what should be learned. 

\subsection{Actionable Learning} A key property of linear control systems is the separation principle. This principle asserts that an optimal estimator can be designed independently from the optimal control. A consequence of the separation principle is that as soon as a measurement has been taken, one knows that the automation system can start to productively take actions. That is, every measurement is \emph{actionable} for the control system. A generalization of the separation principle is to ask whether designing a learning algorithm can be done independently from designing its control system. In general, this review assumes that this is not possible---the learning and control goals are mutually dependent.  However, in some learning cases the relationship between what is being learned and when or how soon one can take action may be important. For instance, in the case of shape recognition in Figure~\ref{fig:shapereconstruction}, exploring an object to determine its shape properties must happen  prior to exploring an unknown environment in search of that shape. This transition is an example of the representation (in this case the abstraction's ``shape'') becoming actionable to the control system. As far as the authors are aware, this topic is little studied in control, but has a long history in psychological study of decision making (\textit{e.g.}, see the many books on this topic by Alain Berthoz~\cite{berthoz2008physiology}).

When a control system becomes actionable is particularly important when distinguishing between active learning and passive learning. During the active learning phase, learning may be the primary goal of the control system. During the passive learning phase the robotic system (or animal) may transition to attempting its ultimate task while continuing to run online passive learning updates. In single-shot learning, where the learner only has one trajectory to exploit for the purpose of learning, being able to robustly detect when learning has become sufficient to take action is a critical part of the path to task success. 

Analysis methods are needed for describing conditions under which learned models are \emph{sufficient} for making a decision to combine the estimation aspects of learning with the control aspects of learning. This transition is often characterized in terms of exploration/exploitation trade-offs~\cite{audibert2009exploration} in the context of sampling-based learning. In the context of a physical system, exploration and exploitation depend on the physics of the learner and environment, and the transition between them will be regulated by the control system. In the case of the example in Figure~\ref{fig:dynamicrecovery}, this would be a safety-critical decision---devoting inadequate time for active learning yields an insufficient model for recovery prior to the vehicle hitting the ground, while engaging in active learning too long will lead to a catastrophic failure. This particular example would likely yield a convex function that represents safety as a function of transition time. However, how to analyze and compute this transition in general is unknown.   

Efficiently forming representations relevant to task completion is part of the challenge in forming actionable representations. When a representation becomes actionable, we capture particular elements of the underlying object or task relevant for decision making while ignoring irrelevant sensory data. The question of determining functionally applicable representations has been explored in~\cite{ghosh2018learning}. The authors claim that the structure of the environment can be modeled with a known goal-conditioned policy---a policy that can achieve a goal state from a given state. The authors refine this policy by differentiating states using the actions necessary to reach them. Thus, states that are functionally similar are closer to each other in the representation than they would be when representing their location with an Euclidean distance. This method could benefit from active learning. For instance, one may use the entropy of the representation rather than the entropy of the input or entropy of the physical states, as the information quantity to force active learning capabilities. However one constructs representations from data-driven experience, an important question will be how to synthesize active learning to close the loop on representation generation.

\section{Conclusion} \label{sec-conclusions}
Active learning and data-driven control will play a major role in future robotic systems operating without access to reliable analytic models or prior data sets in uncertain environments. Robots will need to become fluent learners---routinely investing time and energy in single-shot learning through purposeful data collection and interpretation. This high level goal transcends the capabilities currently available for robotics in machine learning, both in terms of specifying behavior and representing learning goals. Machine intelligence in general has almost entirely been viewed as an extension of estimation theory, focusing on the processing of data. Even reinforcement learning assumes that the data needed for updating a policy is available or that it can be created in simulation. Here we view learning, in part, as an extension of control theory, focusing on how decisions impact learning outcomes. Before these two views can be synthesized into a single coherent theory, many challenges need to be addressed including those mentioned earlier and many not yet understood.

Expanding our notion of a \emph{model} becomes a key effort moving forward. Models should no longer be solely defined by an ordinary differential equation, though ordinary differential equations may still play critical roles during analysis and computation. Instead, a theme in this review is that model-based reasoning needs to admit any set of meta-principles one asserts, such as symmetries in the system, its stability properties, what equilibria are expected, or its logical structure. These assertions will constrain numerical inference, thereby improving learning by reducing the classes of admissible models.

We have outlined and argued for the development of a theory of robot learning---one that deals with the difficulties and constraints that an embodied learning agent would face in the physical world. While much of machine learning has neglected the challenges that physical embodiment brings, this presents a great opportunity for control theorists at-large. The historical arc of robot control has retained a clear focus on the physical properties that ensure safe, robust, and reliable performance. By merging our understanding of controllability, stability, and compliance, with the flexibility of black-box learning, an action-oriented theory of learning will be key to enable future robot technologies.

\begin{center}
    \textbf{Acknowledgements}
\end{center}
We would like to thank Muchen Sun, Ana Pervan, Kyra Rudy, Frank Park, and the anonymous reviewers of the first draft  for their many helpful comments on this manuscript.

\bibliographystyle{elsarticle-num}    
\biboptions{sort&compress}

\bibliography{activelearningbib}

\end{document}